\documentclass{ieeeaccess}
\usepackage{cite}
\usepackage{amsmath,amssymb,amsfonts}
\usepackage{algorithmic}
\usepackage{graphicx}
\usepackage{textcomp}
\def\BibTeX{{\rm B\kern-.05em{\sc i\kern-.025em b}\kern-.08em
    T\kern-.1667em\lower.7ex\hbox{E}\kern-.125emX}}
\usepackage{pifont}       
\usepackage{bbding}       
\usepackage{fontawesome}  
\usepackage{caption}
\usepackage{subcaption}
\usepackage{bbding}
\usepackage{booktabs}

 \usepackage{array}
 \usepackage{multirow}
\begin{document}
\history{Date of publication xxxx 00, 0000, date of current version xxxx 00, 0000.}
\doi{10.1109/ACCESS.2017.DOI}

\title{A Survey on Unsupervised Anomaly Detection Algorithms for Industrial Images}
\author{\uppercase{YAJIE CUI\authorrefmark{}, ZHAOXIANG LIU\authorrefmark{}, and SHIGUO LIAN\authorrefmark{}
},
\IEEEmembership{Member, IEEE}}
\address[]{Unicom Digital Technology, China Unicom, Beijing 100013, China}

\markboth
{Y. Cui \headeretal: A Survey on Unsupervised Anomaly Detection Algorithms for Industrial Images}
{Y. Cui \headeretal: A Survey on Unsupervised Anomaly Detection Algorithms for Industrial Images}

\corresp{Corresponding author: Zhaoxiang Liu (e-mail: liuzx178@chinaunicom.cn), Shiguo Lian (e-mail: liansg@chinaunicom.cn)}
.

\begin{abstract}
In line with the development of Industry 4.0, surface defect detection/anomaly detection becomes a topical subject in the industry field. Improving efficiency as well as saving labor costs has steadily become a matter of great concern in practice, where deep learning-based algorithms perform better than traditional vision inspection methods in recent years. While existing deep learning-based algorithms are biased towards supervised learning, which not only necessitates a huge amount of labeled data and human labor, but also brings about inefficiency and limitations. In contrast, recent research shows that unsupervised learning has great potential in tackling the above disadvantages for visual industrial anomaly detection. In this survey, we summarize current challenges and provide a thorough overview of recently proposed unsupervised algorithms for visual industrial anomaly detection covering five categories, whose innovation points and frameworks are described in detail. Meanwhile, publicly available datasets for industrial anomaly detection are introduced. By comparing different classes of methods, the advantages and disadvantages of anomaly detection algorithms are summarized. Based on the current research framework, we point out the core issue that remains to be resolved and provide further improvement directions. Meanwhile, based on the latest technological trends, we offer insights into future research directions. It is expected to assist both the research community and industry in developing a broader and cross-domain perspective.
\end{abstract}


\titlepgskip=-15pt

\maketitle

\section{Introduction}

\PARstart{I}{ndustry} 4.0 is an era of making use of information technology to promote the industrial revolution, that is, the intelligent era. It is the fourth industrial revolution dominated by intelligent manufacturing. Adhere to the development trend of Industry 4.0, it is the general trend to build a smart manufacture system.

Ideally, once a production link deviates from the standard operation, an alarm signal will be sent, and the producer can make positive improvement response in the shortest time. This transparent and efficient information based production process can minimize production costs and also avoid wasting materials. In the long run, smart manufacturing mode based on artificial intelligence technology \cite{fang2020research} can reduce the requirements on human decision, such as dependence on technical experts, by mining and depositing relevant knowledge, so that labor costs can be saved.

\begin{figure*}[h]%
\centering
    \begin{minipage}[b]{0.18\textwidth}
      \centering
      \includegraphics[width=1\textwidth]{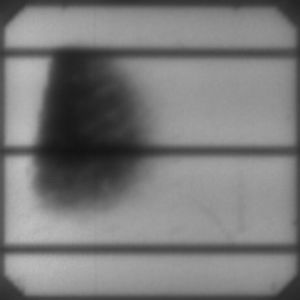}
    \end{minipage}%
    \mbox{\hspace{0.10cm}}
    \begin{minipage}[b]{0.18\textwidth}
      \centering
      \includegraphics[width=1\textwidth]{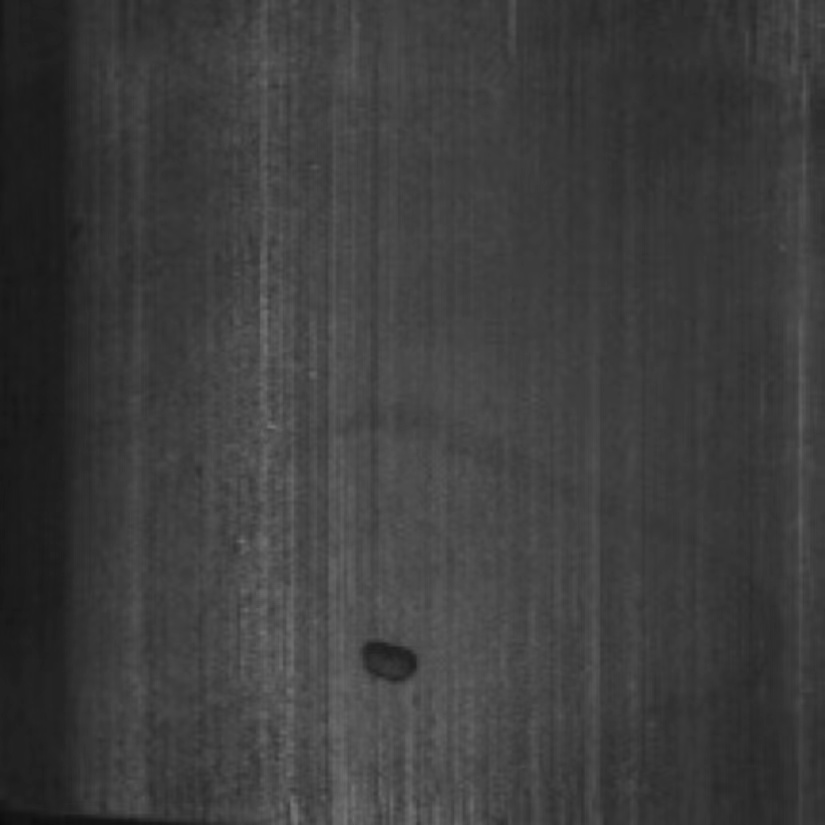}
    \end{minipage}
    \begin{minipage}[b]{0.18\textwidth}
      \centering
      \includegraphics[width=1\textwidth]{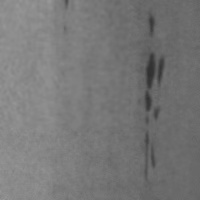}
    \end{minipage}
     \begin{minipage}[b]{0.18\textwidth}
      \centering
      \includegraphics[width=1\textwidth]{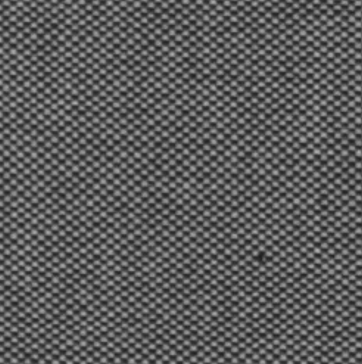}
    \end{minipage}
     \begin{minipage}[b]{0.18\textwidth}
      \centering
      \includegraphics[width=1\textwidth]{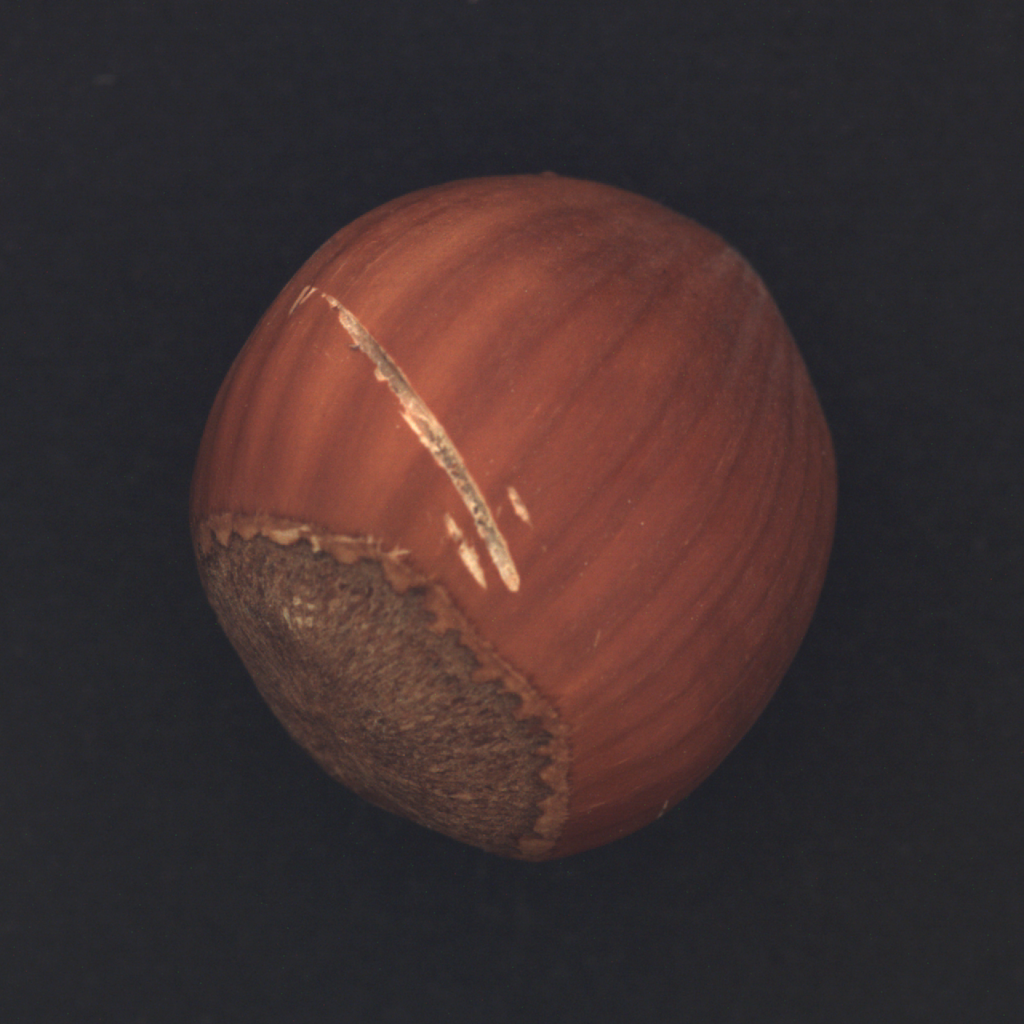}
    \end{minipage}
    
    \begin{minipage}[b]{0.18\textwidth}
      \centering
      \includegraphics[width=1\textwidth]{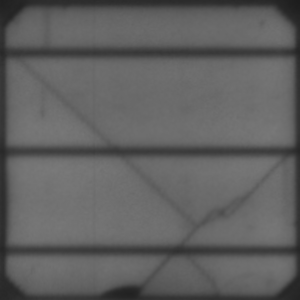}
      \subcaption{ELPV\cite{Deitsch2019,Buerhop2018,Deitsch2021}}
    \end{minipage}%
    \mbox{\hspace{0.10cm}}
    \begin{minipage}[b]{0.18\textwidth}
      \centering
      \includegraphics[width=1\textwidth]{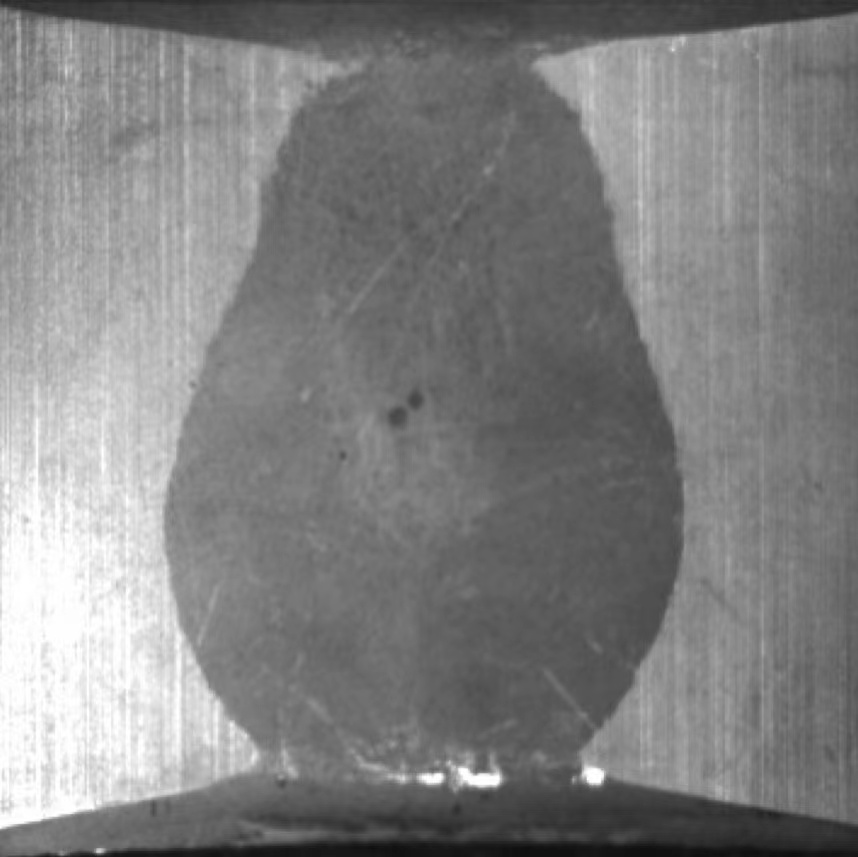}
      \subcaption{NEU-CLS\protect\footnotemark[1]}
    \end{minipage}
    \begin{minipage}[b]{0.18\textwidth}
      \centering
      \includegraphics[width=1\textwidth]{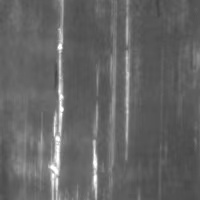}
      \subcaption{MTD\cite{huang2020surface}}
    \end{minipage}
     \begin{minipage}[b]{0.18\textwidth}
      \centering
      \includegraphics[width=1\textwidth]{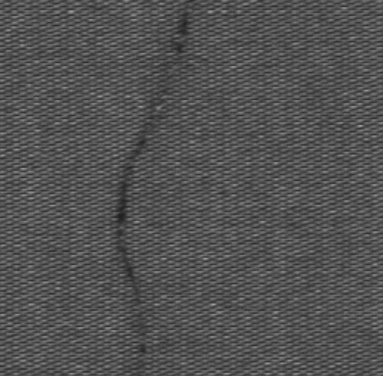}
      \subcaption{AITEX\cite{silvestre2019public}}
    \end{minipage}
     \begin{minipage}[b]{0.18\textwidth}
      \centering
      \includegraphics[width=1\textwidth]{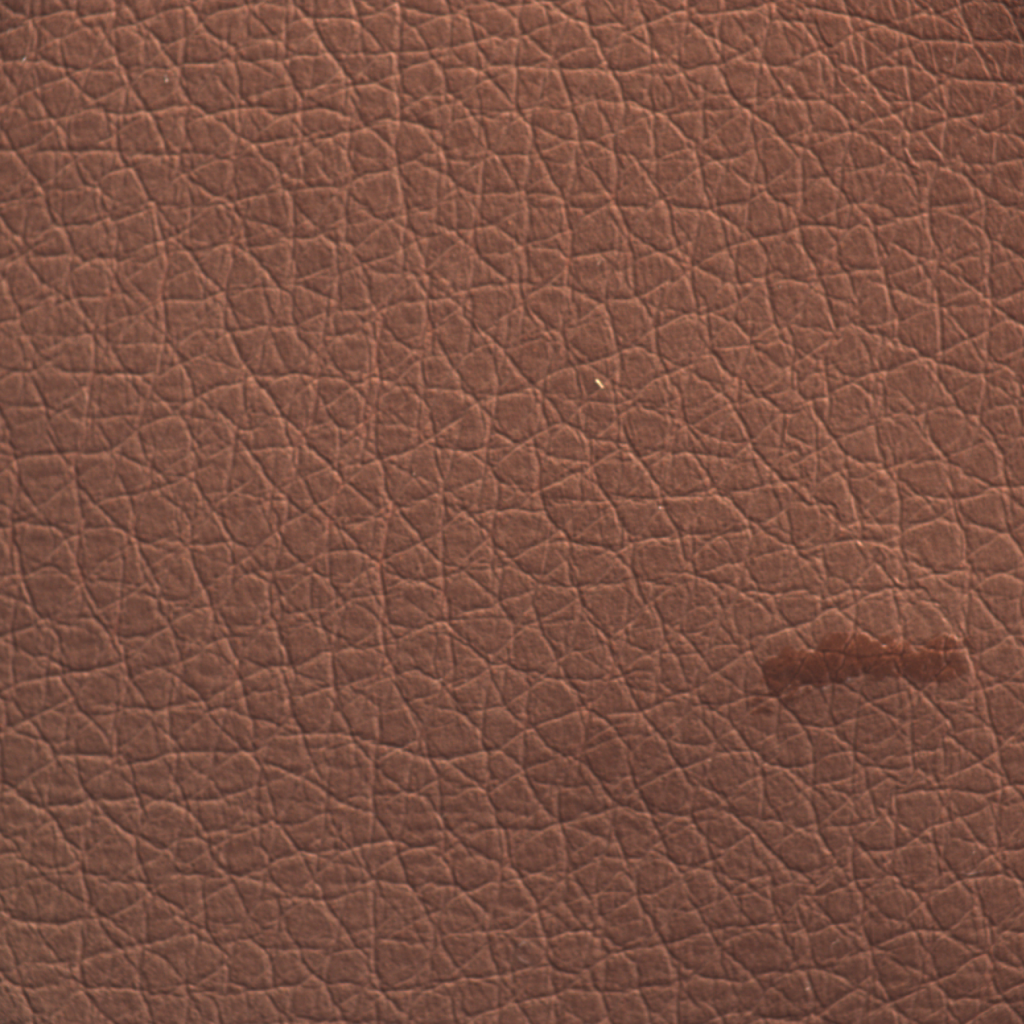}
      \subcaption{MVTec\cite{bergmann2019mvtec}}
    \end{minipage}
   
\caption{Examples of defects of various materials in datasets.}
\label{fig:flaw}
\end{figure*}

Material anomaly extensively exists in industrial production \cite{xie2008review,zhang2020review,fang2020research}, and more and more safety problems are caused by material defects. Hence people pay more attention to the detection of material anomalies. There are a lot of defects images in the industrial scene. Examples of surface defects are shown in Fig.~\ref{fig:flaw}. The traditional method of surface anomaly detection and localization relies on the manual operation by qualified specialists, which is not only inefficient but also depends on the subjective judgment of operators, making it difficult to ensure the accuracy of detection. The production mode of anomaly detection equipment combined with industrial production line ensures the quality of products, reduces the cost of manual testing, and improves the efficiency of production as well. With the rapid rise of computer image processing technology, many algorithms have been gradually applied to the field of material anomaly detection, improving the accuracy of material anomaly detection. In recent years, deep learning \cite{cao2020pixel,pang2021deep} has also been applied to material anomaly detection and achieved extraordinary performance.

\footnotetext[1]{http://faculty.neu.edu.cn/yunhyan/NEU\_surface\_defect\_database.html}

Current deep learning-based supervised algorithms have certain limitations. Model training requires a large amount of labeled data \cite{pang2021deep}, while images with defect labels are not easy to obtain. And the lack of defect samples makes it difficult to bring the models online, which also limits the application of deep learning in the industrial detection field. Therefore, a new solution is urgently needed - unsupervised algorithms, which need no labeled data. This paper provides a review of some recently proposed unsupervised methods, whose innovation points and frameworks are described in detail. Note that we only concentrate on industrial vision anomaly detection algorithms. Particularly, the industry uses the terms defect detection, visual anomaly detection, and surface detection, which we all count in our research. Meanwhile, publicly available datasets for industrial anomaly detection are introduced, with experimental results displayed. This review offers different contributions that distinguish it from other reviews. 
\begin{itemize}
\item[$\bullet$] We discuss the inadequacies of the current algorithms and dataset tailored to the requirements of actual industrial scenarios, such as edible oil impurity detection.
\item[$\bullet$] Based on the current research framework, we suggest that the conflict between FAR (false alarm rate) and MAR (missed alarm rate) is the core issue that remains to be resolved. We also provide further improvement directions to current methods, like integration of diverse technologies.
\item[$\bullet$] We offer insights into future directions based on the latest technological trends, such as foundation model and multimodal learning.
\end{itemize}
 
The subsequent content of the article is organized as follows: related works in Section~\ref{section:Related}, visual anomaly detection methodology in Section~\ref{section:Methodology}, comparison and analysis of methodologies in Section~\ref{section:Analysis}, introduction to industrial datasets in Section~\ref{section:Datasets}, challenge and discussion of research actuality and future development direction in Section~\ref{section:Challenges}, conclusion in Section~\ref{section:Conclusion}.

\section{Related works}
 \label{section:Related}
\noindent\emph{A. Previous Algorithms}

 We only focus on the surprising success and dominance of anomaly detection in industrial images but exclude other areas such as action recognition \cite{wang2021self} \cite{koniusz2021tensor} and video anomaly detection \cite{wang2019loss}. Although some of the strategies have been validated in the above scenario, real industrial images lack prior knowledge of action images and video sequence information, which makes it difficult for models
to generalize across different domains.
 
For some existing methods of visual anomaly detection, the development and changes in technology are introduced below. 
Initially, anomaly detection relied on statistical methods. Statistical approaches assess the geographic distribution of pixel values by extracting statistical information from defect images. Histogram information \cite{chen2009automated,lee2006automated,manish2018machine,chu2017steel,ng2006automatic,aminzadeh2015automatic}, co-occurrence matrices \cite{kumar2018glcm,shabir2019tyre,sadaghiyanfam2018using,deotale2019fabric,yonghua2015study,yang2014automatic}, and local binary patterns (LBP) \cite{sindagi2017domain,song2013noise,fekri2017multi,liu2017fabric,li2019wood,yan2017weld} have all been presented as statistical methods for defect detection. Statistical approaches can present anomalies in an intuitive and discriminative manner, and they are simple to model, interpret, and display. However, they frequently make assumptions, such as separable defect regions, that cannot be satisfied in all scenarios.
Later, hand-extracted features can describe the structure of the image. In structural methods, defect feature is characterized by texture elements \cite{chondronasios2016feature,ma2019algorithm,zhang2013defect,hou2012welding,halim2016automatic,tacstimur2016rail,zhang2018defect,hu2016detection,jayashree2012hybrid,rebhi2016fabric,su2014application}. As a result, the structural approaches' goals are to extract the texture elements of defects, which are used to represent the spatial placement rules. The geometrical feature can be found using structural approaches. This
approach is easier to implement and better suited to random textured defects. However, the majority of them are sensitive to the shape and size of defects, and defect images should be aperiodic.
 
In the field of image processing, filter-dependent methods are also used for anomaly detection. Filter-based methods apply some filter banks on defect images and calculate the energy of the filter responses \cite{ pastor2019surface,bai2014saliency,kumaresan2017defect,ajithaprasad2019defect,hu2015automated,wang2012singular,yang2016defect,liu2017defect,li2015fabric,jing2014supervised,tong2016differential,yang2010multi,vijaykumar2015rail}. Common filter-based methods include Sobel operator, Canny operator, Gabor operator, Laplacian operator, wavelet transform, and Fourier transform, which can be further divided into spatial domain, frequency domain, and spatial-frequency domain methods. In vision-based 
anomaly recognition, filter-based approaches are widespread. The cross-domain methods can aid the model in extracting more meaningful information. Furthermore, they are affine transformation invariant and can handle multi-scale defects. While, they may not be appropriate to random textured images, and some of them may be influenced by feature correlations and noises.


With the development of neural networks and machine learning, a large number of supervised algorithms have appeared. Supervised Neural Networks \cite{li2016deformable,wang2020smart,liu2019research,weimer2016design,li2016automated,kumar2018automated,brackenbury2019automated,sacco2018automated,yang2020high,liu2019real,song2019weak,liu2019periodic}, Support Vector Machines (SVM) \cite{pereira2018goat,tural2019automated,samy2016automatic} and k-Nearest Neighbors \cite{lopez2005surface,lopez2005fast,pernkopf2004detection,mandriota2004filter,wiltschi2000automatic,chan2000fabric,latif2000efficient} are the most common supervised algorithms. Recently, deep learning-based algorithms are becoming popular. The majority of deep learning-based visual anomaly detection is data-driven. To build the visual anomaly detection model, supervised methods take two means. The first one trains an image-level classification model, which requires a labeled training set including both normal and abnormal samples. The second conducts refined object localization. Containing more information, supervised methods should theoretically yield higher detection rates than semi-supervised and unsupervised methods. However, because of lacking training dataset that covers all defect locations and erroneous labeling, there still exists certain technical difficulties. 
 
\noindent\emph{B. Previous Surveys}
\begin{table*}[]
\caption{Summary of previous reviews.}
\label{previous reviews}
\centering
\begin{tabular}{@{}llll@{}}
\hline
\textbf{Tile} &
  \textbf{Year} &
  \textbf{Venue} &
  \textbf{Description} \\ \hline
Deep learning for anomaly detection \cite{Dwang2020deep}&
  2020 &
  KDD &
  \begin{tabular}[c]{@{}l@{}}This review combs \textbf{formulations} of the most \\ classical algorithms of different schools.\end{tabular} \\\hline
\begin{tabular}[c]{@{}l@{}}A review on recent advances in vision-based defect \\ recognition towards industrial intelligence \cite{gao2021review}\end{tabular} &
  2021 &
  \begin{tabular}[c]{@{}l@{}}Journal of \\ Manufacturing \\ Systems\end{tabular} &
  \multirow{7}{*}{\begin{tabular}[c]{@{}l@{}}The reviews present a \textbf{concise overview} of \\ traditional and deep learning-based visual \\ anomaly detection techniques.\end{tabular}} \\
\begin{tabular}[c]{@{}l@{}}Recent advances in surface defect inspection of \\ industrial products using deep learning techniques \cite{zheng2021recent}\end{tabular} &
  2021 &
  \begin{tabular}[c]{@{}l@{}}Advanced \\ Manufacturing \\ Technology\end{tabular} &
   \\
\begin{tabular}[c]{@{}l@{}}Visual Anomaly Detection for Images: A Systematic \\ Survey\cite{yang2022visual}\end{tabular}                                      & 2022 & \begin{tabular}[c]{@{}l@{}}Procedia \\ Computer \\ Science\end{tabular} &     \\\hline
\begin{tabular}[c]{@{}l@{}}Visual-based defect detection and classification \\ approaches for industrial applications—a survey \cite{czimmermann2020visual}\end{tabular} &
  2020 &
  Sensors &
  \\
\begin{tabular}[c]{@{}l@{}}A review on industrial surface defect detection \\ based on deep learning technology \cite{qi2020review}\end{tabular} &
  2020 &
  MLMI & \multirow{5}{*}{\begin{tabular}[c]{@{}l@{}}These studies focus only on anomaly \\ detection on \textbf{particular materials}.\end{tabular}}
   \\
\begin{tabular}[c]{@{}l@{}}Fabric defect detection using computer vision \\ techniques: A comprehensive review \cite{rasheed2020fabric}\end{tabular} &
  2020 &
  \begin{tabular}[c]{@{}l@{}}Mathematical\\ Problems in \\ Engineering\end{tabular} &
   \\
\begin{tabular}[c]{@{}l@{}}Research progress of automated visual surface \\ defect detection for industrial metal planar \\ materials \cite{fang2020research}\end{tabular} &
  2020 &
  Sensors &
   \\\hline
\begin{tabular}[c]{@{}l@{}}Image-based surface defect detection using deep \\ learning: A review \cite{bhatt2021image}\end{tabular} &
  2021 &
  \begin{tabular}[c]{@{}l@{}}Computing and \\ Information \\ Science in \\ Engineering\end{tabular} &
  \begin{tabular}[c]{@{}l@{}}For the unsupervised aspect, more attention \\ is paid to the different \textbf{network} \\ \textbf{architectures} used by different types of \\ algorithms.\end{tabular} \\\hline
\begin{tabular}[c]{@{}l@{}}A unified survey on anomaly, novelty, open-set, and \\ out-of-distribution detection: Solutions and future \\ challenges \cite{salehi2021unified}\end{tabular} &
  2021 &
  arXiv &
  \begin{tabular}[c]{@{}l@{}}The article focuses on the \textbf{intersection} of \\ different research fields, de-emphasizing \\ methodological schools and blurring domain \\ boundaries.\end{tabular} \\ \hline
\begin{tabular}[c]{@{}l@{}}GAN-based Anomaly Detection: A Review\cite{xia2022gan}\end{tabular} & 2022& Neurocomputing & \begin{tabular}[c]{@{}l@{}}This paper focus on \textbf{GAN-based} anomaly  \\ detection and discusses its theoretical basis \\and applications.\end{tabular}  \\ \hline
\end{tabular}\\

\end{table*}

As an important research area, there have been already many reviews researching it. Table.~\ref{previous reviews} enlists some existing surveys in the industrial anomaly detection field, which have different focus from our research.
Specifically, the review \cite{Dwang2020deep} starts with formulations of the most classical algorithms of different schools. 
The paper \cite{gao2021review} surveys traditional methods, and also introduces deep learning-based methods. For each type of method, the characteristics of each method are listed in a general way. The article \cite{zheng2021recent} represents an introduction to traditional methods, deep learning-based methods, and an introduction to hardware and software devices is involved in the meantime. For the introduction to deep learning-based methods, it mainly focuses on the supervised domain and introduces milestone algorithms by timeline. 
The survey \cite{yang2022visual} groups the relevant approaches given their underlying principles and discusses their assumptions, advantages, and disadvantages.
The study \cite{czimmermann2020visual} focuses on specific solutions for visual processing methods and, in particular, visual inspection approaches for metallic, ceramic, and textile surfaces in industrial applications. The methods in literature \cite{qi2020review} are divided into categories based on the types of detection materials used. Some studies focus only on anomaly detection on a particular material. A thorough survey \cite{fang2020research} is provided of both two-dimensional and three-dimensional surface defect detection systems for various common metal planar material products such as steel, aluminum, copper plates, and strips. The review \cite{rasheed2020fabric} presents a detailed overview of histogram-based approaches, color-based approaches, image segmentation-based approaches, frequency domain operations, texture-based defect detection, sparse feature-based operations, and image morphology operations for fabric defect detection. 
The article \cite{bhatt2021image} investigates supervised and semi-supervised deep learning algorithms. As for the unsupervised aspect, more attention is paid to the different network architectures used by different types of algorithms. 
The article \cite{salehi2021unified} focuses on the intersection of different research fields, providing extended cross-cutting ideas, exhaustively introducing the algorithms and frameworks of some typical methods, de-emphasizing methodological schools and blurring domain boundaries. It intends to bring these fields closer together.
The article \cite{xia2022gan} focuses only on GAN-based algorithms.

In practice, it is more biased towards the needs for the unsupervised domain in the current industrial context. To our best knowledge, no review has been done for the recently emerged unsupervised methods. The article will provide a comprehensive and in-depth summary of the state-of-the-art algorithms for visual industrial anomaly detection, which will be divided into a systematic categorization listed as ~\ref{section:Reconstruction} Reconstruction-based, ~\ref{section:nf} Normalizing Flow (NF)-based,
~\ref{section:Representation} Representation-based, 
~\ref{section:augmentation} Data augmentation-based, and ~\ref{section:enhancements} Algorithm enhancements. 
This comprehensive summary is expected to contribute to the implementation and practice of the industrial field.

\section{Methodology}
 \label{section:Methodology}
A large part of the traditional visual anomaly detection algorithms belong to the category of supervised learning \cite{gornitz2013toward,kawachi2018complementary}, which requires collecting enough samples of different defect categories and accurate labeling, such as the category of the image, the location of the defects in the image and the category information of each pixel. However, in many application scenarios, it is difficult to collect a sufficient number of samples \cite{ruff2021unifying}. For example, in the surface defect detection task, most of the images collected actually belong to normal defect-free samples, while only a small amount belong to defect samples. With diverse types of defects to be detected, the number of defect samples available for training is very limited.

\begin{figure}[h]%
\begin{minipage}[t]{\linewidth} 
\centering 
    \includegraphics[width=1\textwidth]{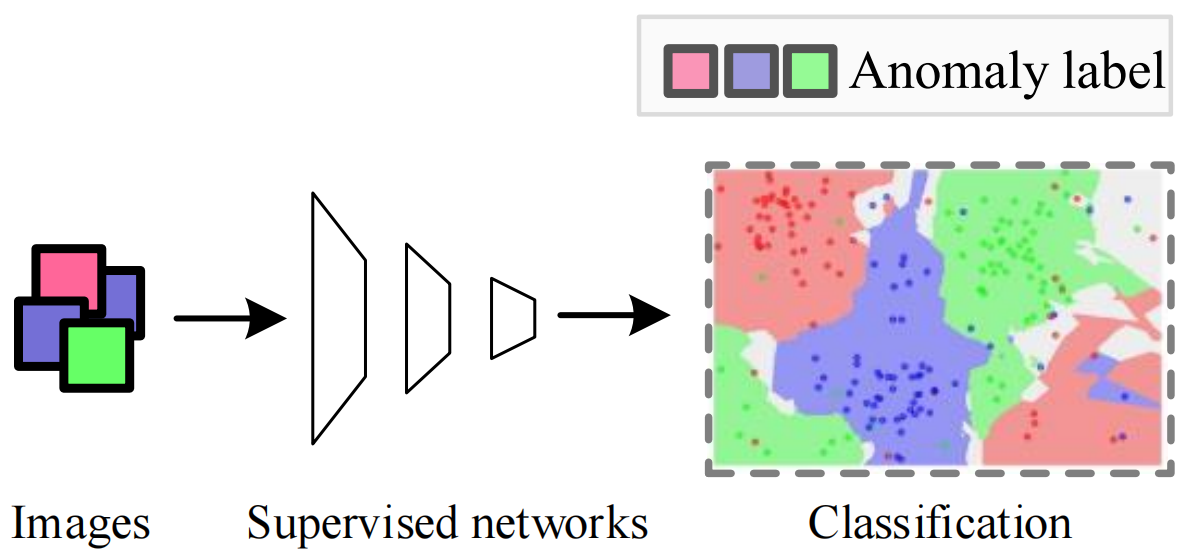}
    \subcaption{Supervised} 
    \label{frame} 
    \end{minipage}%
    
    \begin{minipage}[t]{\linewidth} 
    \centering 
    \includegraphics[width=1\textwidth]{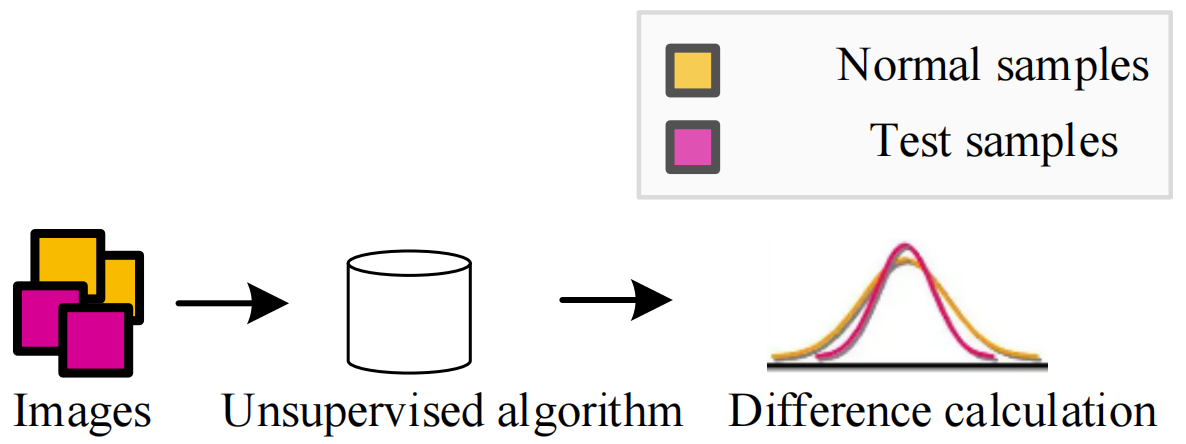}
    
    \subcaption{Unsupervised} 
    \end{minipage} 
    
    \caption{Comparison of framework diagrams of supervised and unsupervised algorithms.} 
    \label{fig:supervise and unsupervise framework} 
\end{figure}

Unsupervised visual anomaly detection algorithms can build detection models without any annotated samples, which makes it very suitable for the scenarios described above. In anomaly detection tasks, the difficulty in collecting normal images is much lower than that of anomalous images, which can significantly reduce the time and labor cost of detection algorithms in practical applications. Moreover, unsupervised visual anomaly detection models detect anomalous samples by analyzing the differences between normal samples and abnormal samples, allowing the algorithm to detect a wide range of abnormal samples, even brand new sorts of flaws. Comparison of framework diagrams of supervised and unsupervised algorithms is shown in Fig.~\ref{fig:supervise and unsupervise framework}.

There are some highlighted approaches worth mentioning among the algorithms with great performance, which will be described in detail in this section. We categorize the existing research into five types: reconstruction-based methods, normalizing flow-based methods, representation-based methods, data augmentation-based methods, and algorithm enhancement. An overview and summary of these categories is listed in Table~\ref{tab:overview and summary}.

\begin{table*}[]
\caption{An overview and summary of different categories of anomaly detection algorithms.}
\label{tab:overview and summary}
\begin{tabular}{@{}p{2cm}p{4cm}p{3cm}p{3cm}p{3cm}@{}}
\hline
\multicolumn{1}{c}{Category} &
  \multicolumn{1}{c}{Description} &
  \multicolumn{1}{c}{Paper} &
  \multicolumn{1}{c}{Advantages} &
  \multicolumn{1}{c}{Disadvantages} \\ \hline
Reconstruction-based Methods &
  The core idea is to conduct encoding and decoding on the input normal images and train the neural network with the aim of reconstruction. Based on the assumption that by training only on normal images, the model will not be able to reconstruct abnormal images correctly, and the anomaly scores will be higher. &
  Ven.,CAVGA\cite{venkataramanan2020attention}; Liu,UTAD\cite{liu2021unsupervised}; Yang,DFR\cite{yang2020dfr}; Mo.,STPM\cite{STPM}; Yamada,RSTPM\cite{rstpm}; Deng,RDOE\cite{deng2022anomaly}; Rudolph,AST\cite{rudolph2023asymmetric}; Massoli,MOCCA\cite{massoli2020mocca}; Liang,OCR-GAN\cite{liang2022omni}; Mishra,VT-ADL\cite{MishraVFPF21}; Li,SOMAD\cite{li2021anomaly} &
  The algorithm principle is straightforward and comprehensible, while the network architecture is uncomplicated. &
  Reconstructors like auto-encoders and GANs are highly generalizable and robust, and the anomalous part of the image can be well reconstructed, leading to hypothesis failure. \\\hline
  
Normalizing Flow (NF)-based Methods &
  NF is able to learn transformations between data distributions and well-defined probability density functions, which can serve as a suitable estimator of probability densities for the purpose of detecting anomalies. &
  Rudolph,Differnet\cite{rudolph2021same}; Gudovskiy,CFlow\cite{gudovskiy2022cflow}; Rudolph,FCCSF\cite{rudolph2022fully}; Yu,FastFlow \cite{yu2021fastflow} &
  NF mapping is bijective and can be evaluated in both directions and inference efficiency is high. &
  NF-based methods require expensive training computational resources. \\\hline
  
Representation-based Methods &
  Deep neural networks are used to extract meaningful vectors describing the image, and the anomaly score is usually represented by distance calculation. &
  Cohen,SPADE\cite{cohen2020sub}; Defard,PaDIM \cite{defard2021padim}; Kars.,PatchCore \cite{roth2021towards}; Wang,GP\cite{wang2021glancing}; Zheng,FYD \cite{zheng2021focus}; Rip.,Gaussian-AD\cite{rippel2021modeling}; Yi,Patch SVDD\cite{yi2020patch}; DisAug CLR\cite{sohn2020learning}; Ki.,Semi-orth\cite{kim2021semi}; Lee,CFA\cite{lee2022cfa} &
  Representation-based methods do not call for a dedicated training stage, which introduces no parameters other than the backbone. &
  Because the backbone is usually biased towards ImageNet, it does not have good generalizability for different domains. \\\hline
  
Data augmentation-based Methods &
  Augmentation algorithms are designed to resemble anomalies. &
  Zav.,DRAEM\cite{zavrtanik2021draem}; Schluter,NSA\cite{schluter2021self}; Li,CutPaste\cite{li2021cutpaste}; Nicolae,SSPCAB\cite{ristea2021self}; Song,AnoSeg \cite{song2021anoseg} &
  The approach is uncomplicated and facile to comprehend and execute. &
  The data augmentation method is unable to fully replicate actual anomalies, resulting in certain generalization issues. \\ \hline
\end{tabular}
\end{table*}
\subsection{Reconstruction-based Methods}
 \label{section:Reconstruction}
\begin{figure}[h]%
\centering
\includegraphics[width=0.45\textwidth]{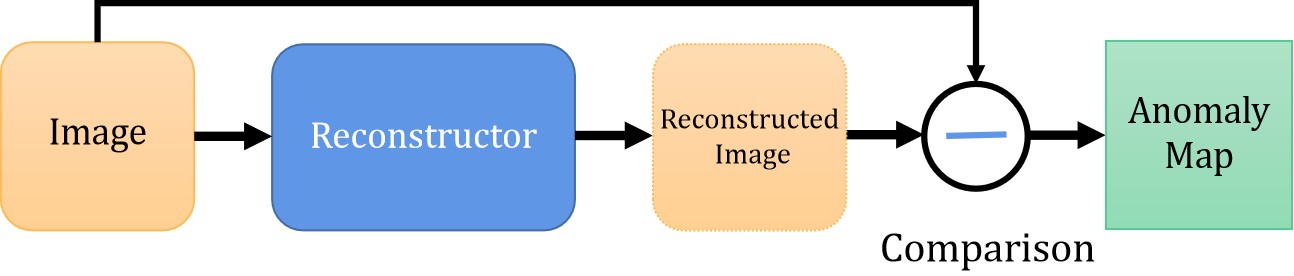}
\caption{The basic flow of reconstruction-based method. The image is input into a reconstructor such as AE and GAN, to output the reconstructed image and compare it with the original input image. The difference between the two is used to get an anomaly map so as to realize anomaly detection.
 }\label{fig:reconstruction}
\end{figure}

To learn the distribution pattern of normal images, the core idea is to conduct encoding and decoding on the input normal images and train the neural network with the aim of reconstruction. With the help of the trained networks, the differences between the images before and after reconstruction are analyzed to detect anomalies in the detection stage. With anomaly score usually represented by reconstruction error, the anomalous images are easy to be found because they cannot be reconstructed well. Classical methods based on reconstruction include autoencoders (AE \cite{lecun1989generalization,RudolphWR19})  , variational auto-encoders (VAE \cite{autoencoder2014}) and generative adversarial networks (GAN \cite{goodfellowPMXWOCB14}), which can generate samples from the manifold of the training data. During the training phase, only normal data without anomalies are conventionally modeled. In testing phase, anomaly scores are calculated with the difference between the input image and the reconstructed image. Based on the assumption that by training only on normal images, the model will not be able to reconstruct abnormal images correctly, and the anomaly scores will be higher. The basic flow of reconstruction-based method is shown in Fig.~\ref{fig:reconstruction}.  
\begin{figure*}[h]%
\centering
\includegraphics[width=0.8\textwidth]{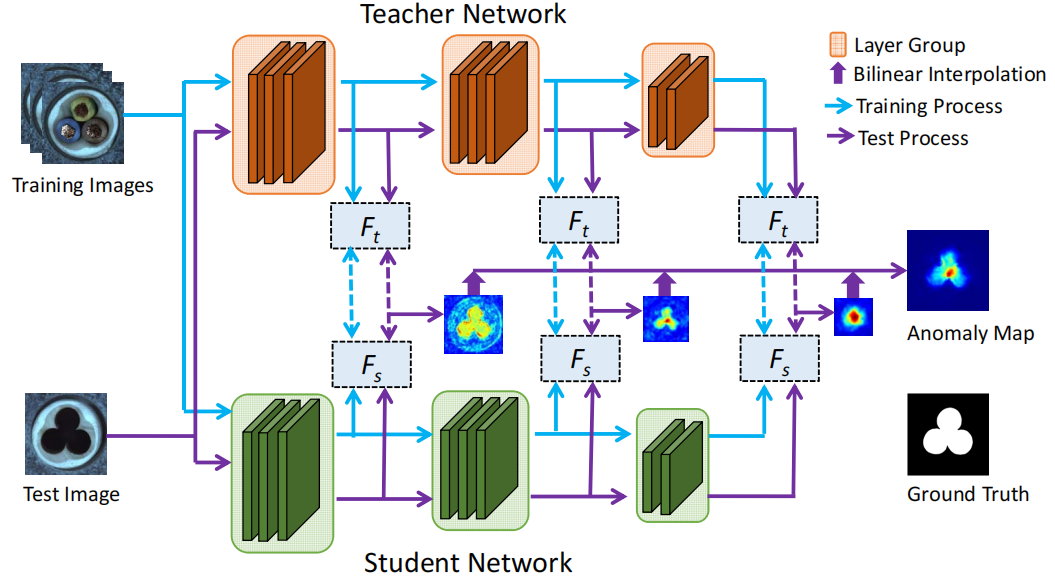}
\caption{Schematic overview of STPM \cite{STPM}. A student network's feature pyramid is trained to match the counterpart of a pre-trained teacher network. If the features from the two models disagree significantly, a test image (or pixel) gets a high anomaly score. STPM approach can detect anomalies of various sizes with a single forward pass owing to the feature pyramid matching scheme. }\label{fig:stpm}
\end{figure*}
Typical auto-encoder and GAN based anomaly detection approaches mentioned above have some limitations, including:

\noindent\textbf{Uncertain threshold.} Autoencoders and GAN based approaches use a thresholded pixel-wise difference between the input and reconstructed image to localize anomalies. However, the use of anomalous training images, which may not be available in real-world situations, is required for these approaches to determine class-specific thresholds. 

\noindent\textbf{High computational cost.} Autoencoder and GAN based anomaly detection approaches often require large amounts of computational resources to train and evaluate. This can be a bottleneck for real-time applications that require fast anomaly detection.

\noindent\textbf{Difficult to interpret.} The representations learned by autoencoders and GANs may be difficult to interpret, making it challenging to understand why a particular instance was classified as an anomaly. This can make it difficult to diagnose and fix problems in the system.

\noindent\textbf{Sensitivity to hyperparameters.} The performance of autoencoder and GAN based anomaly detection approaches can be sensitive to the choice of hyperparameters, such as the number of layers, the learning rate, and the batch size. It can be challenging to select the optimal hyperparameters for a given dataset, and the performance can degrade significantly if the hyperparameters are not tuned properly.

To improve the reconstruction ability, in method CAVGA \cite{venkataramanan2020attention}, VAE, GAN and other means are combined and an attention mechanism is introduced for the first time into the anomaly detection field. The framework encourages the attention map to cover the entire normal region, while suppressing attention maps corresponding to the anomaly classes in the training images. Two modes of unsupervised and weakly supervised are provided. 1. Unsupervised mode: GAN is used as the overall architecture, VAE is used as the codec and attention map is generated by Grad-CAM. Loss function consists of three parts: VAE, adversarial loss, and attention part. 2. Weakly supervised mode: Compared with mode 1, classifiers are added to distinguish normal and abnormal samples. Loss function consists of four parts: VAE, adversarial loss, complementary guided attention loss, and classification loss. However, one potential drawback is that it relies on the assumption that anomalies in images are always visually distinct from the background or normal regions. This may not always be the case, as certain types of anomalies may be visually similar to the background, making them harder to detect using this method. 

Classical methods like GAN and Autoencoder compare the input and its difference from the output to pinpoint the anomaly. However, coarse reconstructions produce excessive image differences, which prevents the detection of anomalies. To address this problem, the approach UTAD \cite{liu2021unsupervised} proposes an unsupervised visual anomaly detection method for natural images by combining mutual information, GAN, and autoencoder. A two-stage framework (i.e., IE-Net, Expert Net) is utilized to generate high-fidelity and anomaly-free input reconstructions for anomaly detection tasks.

Aiming at the anomalies in small and confined regions of images, DFR method \cite{yang2020dfr} suggests an effective unsupervised anomaly segmentation approach, which utilizes the transformed hierarchical CNN features to build dense discriminative multiscale feature representations for every local region of the images via a specially designed regional feature generator. DFR also proposes to detect possible anomalous regions in images by deep feature reconstruction, i.e. reconstructing the multiscale regional features via a deep yet efficient convolutional autoencoder(CAE). The regional feature generator takes the multi-scale feature maps as input and turns them into a relatively large single feature map, which is then reconstructed by a deep CAE. By calculating the reconstruction error and the anomaly score map, anomalies will be segmented if any score on the anomaly map is greater than the estimated value or a user-defined threshold.

Some attempts utilize pre-trained model of image classification task. Nevertheless, the problem of the incompleteness of transferred knowledge and the complexity of handling scaling has not yet been resolved. Thus, STPM \cite{STPM} introduces a novel feature pyramid matching technique and incorporates it into the student-teacher anomaly detection framework. Fig.~\ref{fig:stpm} shows the overview of STPM. The algorithm employs multiple layers of features extracted from a powerful network pre-trained for image classification tasks as the teacher to guide a student network with the same structure to learn the distribution of anomaly-free images. The student network learns the distribution of images by matching the features of the anomaly-free images with the pre-trained network, and this step of transmission seeks to retain as much critical information as possible. In the training phase, the teacher network is a mature network trained on ImageNet, and the image input network generates multi-layer feature maps. The student network is trained with a fraction of the training set, approximating the multi-layer feature trained by the teacher network as much as possible. In the testing phase, samples are put into both teacher and student networks, the differential loss between which is computed. A high anomaly score will be assigned if the features of a test image (or pixel) deviate significantly between the two models. If any pixel in the image is anomalous, the image is judged as anomalous.

The RSTPM \cite{rstpm} approach is a generalization of the Student-Teacher framework method STPM, which was developed previously. The new approach differs from prior STPM in three ways: student network for reconstruction, an attention mechanism from the teacher network to the student network, and a different teacher network structure from the original STPM. 

T-S model typically uses similar or identical architectures. To improve the T-S model’s representation diversity on unknown, out-of-distribution samples, a novel T-S model \cite{deng2022anomaly} with a teacher encoder and student decoder is suggested, along with a straightforward yet powerful reverse distillation paradigm. Instead of receiving raw images directly, the student network takes the teacher model’s one-class embedding as input and targets to restore the teacher’s multiscale representations. It is the first approach to adopt an encoder and a decoder to construct the T-S model. This strategy differs from existing ones due to the heterogeneity of the teacher and student networks and reversed data flow in knowledge distillation.

Previous methods suffer from the similarity of student and teacher architecture, such that the distance is undesirably small for anomalies. To tackle this problem, AST \cite{rudolph2023asymmetric} proposes asymmetric student-teacher networks, which train a normalizing flow for density estimation as a teacher and a conventional feed-forward network as a student to trigger large distances for anomalies.
 
Explicitly leveraging the networks’ multi-layer composition, MOCCA \cite{massoli2020mocca} exploits the output of a deep model at different depths to detect anomalous input in the one-class setting. With MOCCA, the training technique is split into two stages in which the autoencoder is trained on the reconstruction task only, and then only the encoder is utilized to detect anomalies by exploiting a one-class-like objective applied to different layers of the network.

OCR-GAN \cite{liang2022omni} reconsiders the distinction between normal and abnormal images from the frequency domain perspective and proposes a novel framework for anomaly detection based on omni-frequency reconstruction. Specifically, FD module is proposed to decouple the input image into various frequencies and model the reconstruction process as a combination of parallel omni-frequency image restorations.

VT-ADL \cite{MishraVFPF21} combines the classic reconstruction-based methods with the benefits of a patch-based approach. Visual transformer networks contribute to preserving the spatial information of the embedded patches, which is later coped with a Gaussian mixture density network to localize the anomalous areas.

Self-organizing map for anomaly detection (SOMAD) \cite{li2021anomaly} makes use of pre-trained CNN to extract the features of patches and leverages the SOM to maintain the neighborhood relationship of embedding vectors in topology space. It greatly reduces the search space by mapping the normal feature space into 2-dimensional space through SOM. 

\subsection{Normalizing Flow (NF)-based Methods}
 \label{section:nf}

Normalizing Flows (NF) \cite{rezende2015variational} are neural networks that are able to learn transformations between data distributions and well-defined probability density functions. Their special property is that their mapping is bijective and they can be evaluated in both directions. The property of normalizing flows to serve as a suitable estimator of probability densities for the purpose of detecting anomalies has not drawn much attention yet. Here we summarize some recently emerged NF-based visual anomaly detection algorithms to provide ideas for future study.

\begin{figure}[h]%
\centering
\includegraphics[width=0.45\textwidth]{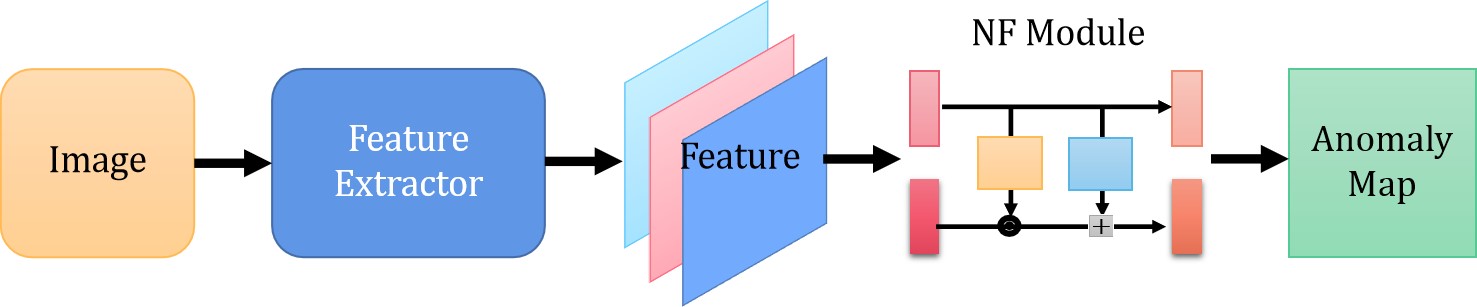}
\caption{The basic flow of the Normalizing Flow (NF)-based method. The features are first extracted by a feature extractor and then fed into NF module to estimate the probability density.  The probability value is employed as the anomaly score in testing phase.}\label{fig:nf}
\end{figure}

There are methods \cite{rudolph2021same,gudovskiy2022cflow} adopting normalizing flow to estimate distribution through a trainable process that maximizes the log-likelihood of normal image features. Normal image features are embedded into standard normal distribution and the probability is used to identify and locate anomalies. The basic flow of the Normalizing Flow (NF)-based method is shown in Fig.~\ref{fig:nf}. 

The method \cite{rudolph2022fully} detects and locates defects based on density estimates of feature maps extracted from input images of different sizes. Cross-connections between scales are made by jointly processing multiscale feature maps using a fully convolutional normalizing flow. 

However, in order to estimate the distribution, the original one-dimensional normalizing flow model must flatten the two-dimensional input feature into a one-dimensional vector, which destroys the inherent spatial positional relationship implied by the two-dimensional image and constrains the NF model. FastFlow \cite{yu2021fastflow} expands the original normalizing flow model to two-dimensional space to address the concerns mentioned above.
\begin{figure*}[h]%
\centering
\includegraphics[width=1.0\textwidth]{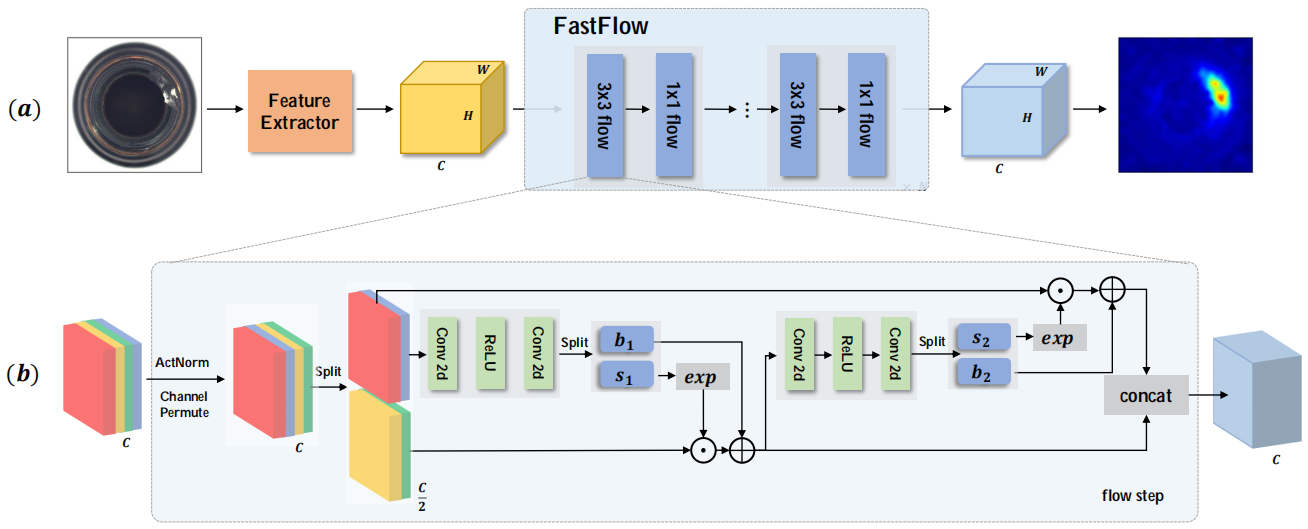}
\caption{(a) shows the whole framework of FastFlow \cite{yu2021fastflow} algorithm. (b) is one flow step of FastFlow. }\label{fig:fastflow}
\end{figure*}
As shown in Fig.~\ref{fig:fastflow}, the algorithm is summarized: the visual features are first extracted by a feature extractor and then fed into FastFlow to estimate the probability density. In the training phase, FastFlow is trained with normal images to transform the original distribution into a standard normal distribution in a 2D manner. In inference phase, the probability value of each position on the 2D feature is employed as the anomaly score.

\subsection{Representation-based Methods}
 \label{section:Representation}

\begin{figure}[h]%
\centering
\includegraphics[width=0.45\textwidth]{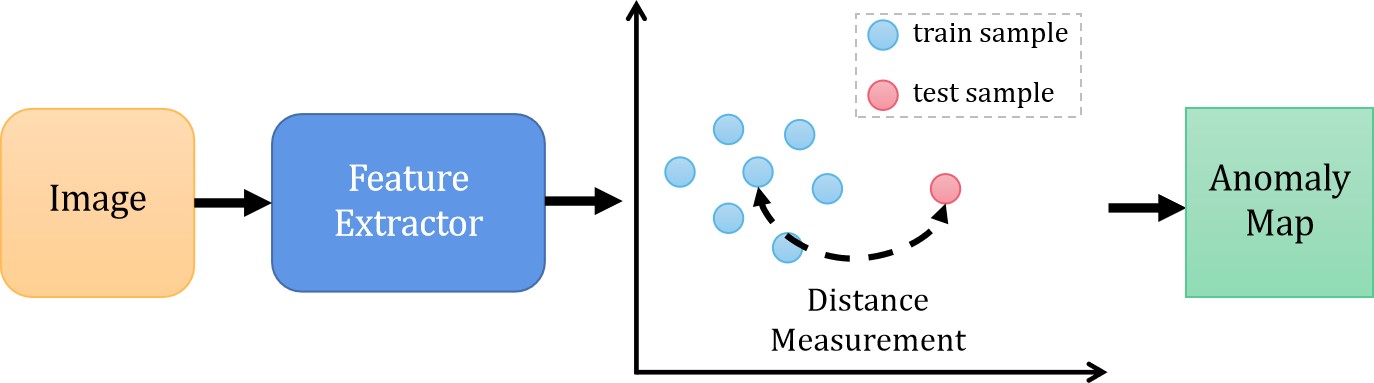}
\caption{The basic flow of representation-based method. Pre-trained deep neural networks are used as feature extractor to extract meaningful vectors describing the input image, and the anomaly map is usually represented by the distance between the test embedded vectors and the reference vector representing normality from the training dataset.}\label{fig:representation}
\end{figure}

For representation-based methodology, deep neural networks are used to extract meaningful vectors describing the entire image, and the anomaly score is usually represented by the distance between the embedded vectors of the test images and the reference vector representing normality from the training dataset. The basic flow of representation-based method is shown in Fig.~\ref{fig:representation}. The core idea is to train a deep neural network as a feature extractor to make the distribution of feature vectors extracted from normal images as compact as possible, i.e., the intra-class distance of the samples is reduced as much as possible. Contrary to reconstruction-based algorithms, representation-based methods do not call for a dedicated training stage, which introduces no parameters other than the backbone. The concept of distance metric learning techniques is comparable to clustering.

\begin{figure*}[h]%
\centering
\includegraphics[width=1.0\textwidth]{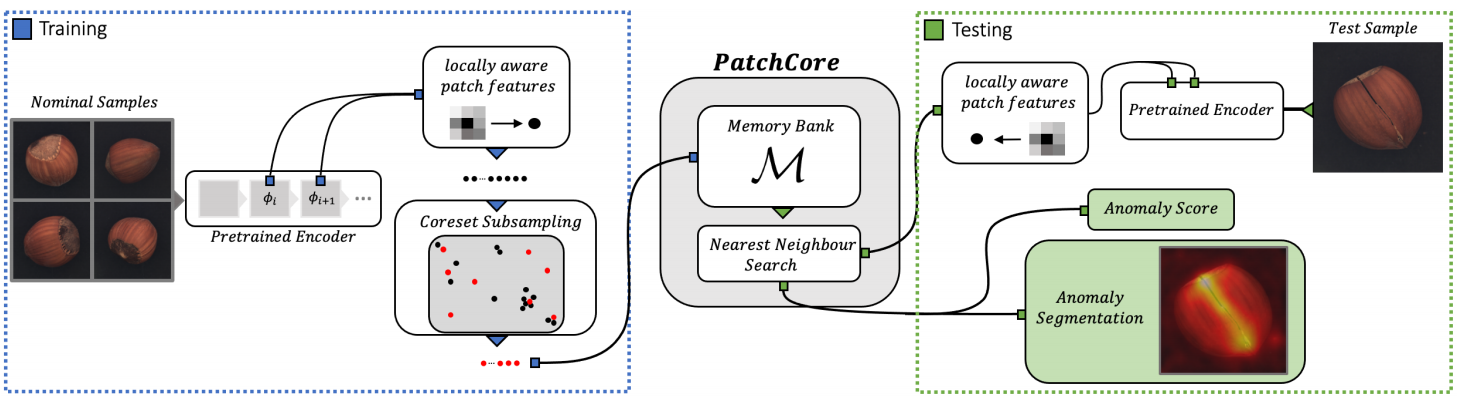}
\caption{The overall framework of PatchCore. During training, normal samples are decomposed into a memory bank of neighborhood-aware patch-level features. To reduce redundancy and inference time,  this memory bank is downsampled via greedy coreset subsampling algorithm. During test time, images are classified as anomalies if at least one patch is anomalous, and pixel-level anomaly segmentation is generated by scoring each patch feature. }\label{fig:patchcore}
\end{figure*}

In the testing phase, most methods calculate the distance between the features of the sample to be tested and the normal features as a metric to perform anomaly detection. Typical algorithms mainly include SPADE \cite{cohen2020sub}, PaDIM \cite{defard2021padim}, PatchCore \cite{roth2021towards}, GP \cite{wang2021glancing}, etc. To record anomaly score and generate a score map, all these approaches employ different distance measurements (loss functions). 
PatchCore uses a maximally representative memory bank of nominal patch-features to integrate embeddings from ImageNet models with an outlier detection model. The framework of PatchCore \cite{roth2021towards} is shown in Fig.~\ref{fig:patchcore}

PaDiM makes use of a pre-trained convolutional neural network (CNN) for patch embedding, and multivariate Gaussian distributions to get a probabilistic representation of the normal class. It also exploits the correlations between different semantic levels of the CNN to better locate the anomalies.

Based on alignment between an anomalous image and a constant number of similar normal images, SPADE \cite{cohen2020sub} uses KNN and multiscale feature pyramid for defect detection and localization of anomalies. The following steps make up SPADE algorithm: i) image feature extraction ii) K-nearest-neighbor normal image retrieval iii) pixel alignment with deep feature pyramid correspondences. 

With a coarse-to-fine alignment technique, FYD method \cite{zheng2021focus} seeks to learn dense and compact distribution of normal images. In both picture and feature levels, the coarse alignment stage normalizes the pixel-level position of objects. After that, the fine alignment stage maximizes the similarity of features across all corresponding locations in a batch.

In terms of the scale of image processing, the methods can be divided into image level, patch level and pixel level. Gaussian-AD \cite{rippel2021modeling} extracts discriminative feature vectors from normal images. Algorithms like Patch SVDD \cite{yi2020patch}, PatchCore \cite{roth2021towards} and PaDIM \cite{defard2021padim} extract discriminative feature vectors from normal image patches. SPADE \cite{cohen2020sub} extracts discriminative features which are used for pixel-level image alignment. 
From different process levels, these methods extract features of normal images and model the distribution with statistical methods. Based on the assumption that abnormal samples have different distributions, more promising results for anomaly detection are yielded.

Based on distribution-augmented contrastive learning, DisAug CLR algorithm \cite{sohn2020learning} first learns self-supervised representations from one-class data, and then builds one-class classifiers on learned representations.

The method Semi-orthogonal \cite{kim2021semi} is a generalization of the prior work's random feature selection method PaDIM \cite{defard2021padim}. It extends the random feature selection to semi-orthogonal embedding as a low-rank approximation of precision matrix for the Mahalanobis distance.

Albeit simple and efficient, most of these methods require manual specification of feature centers in advance, and additional tasks need to be designed in the training stage to avoid model degradation. The approach of setting only one global feature centroid imposes some constraints on the image context. In the changeable scenes of medical images or natural images, it may be challenging to map all images to the same target point under the condition of guaranteeing generalization ability.

 Previous studies focused on approximating the distribution or extracting features with pre-trained CNNs of normal data, which may make the normality of abnormal features overestimated. CFA \cite{lee2022cfa} performs transfer
learning on the target dataset as a solution to alleviate this problem. CFA first acquires multiscale feature maps with biased CNN to generate a patch memory bank. Through transfer learning and the feature adaptation of patch descriptor associated with the memory bank, CFA achieved successful target-oriented anomaly detection.

\subsection{Data augmentation-based Methods}
 \label{section:augmentation}
\begin{figure}[h]%
\centering
\includegraphics[width=0.45\textwidth]{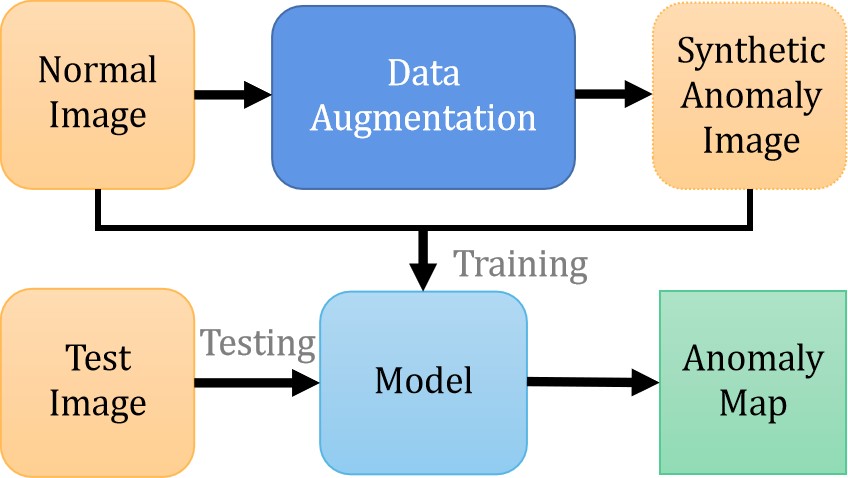}
\caption{The basic flow of the data augmentation-based method. The normal image is augmented to obtain synthetic abnormal image. Both normal samples and synthetic abnormal samples are input into the model for training. The difference generated after the test sample input into the model is regarded as anomaly map.} \label{fig:augmentation}
\end{figure}

In the unsupervised setting, the training data are all anomaly-free data. Hence, there are some algorithms \cite{sohn2020learning,zavrtanik2021draem,schluter2021self,li2021cutpaste,song2021anoseg} that adopt the method of creating anomalies. To
overcome the limitation of insufficient data, augmentation algorithms \cite{Wang2020ASO,Ishida2020UnsupervisedAD} have been widely used in deep learning scheme. The basic flow of the data augmentation-based method is shown in Fig.~\ref{fig:augmentation}.  

\begin{figure*}[h]%
\centering
\includegraphics[width=1.0\textwidth]{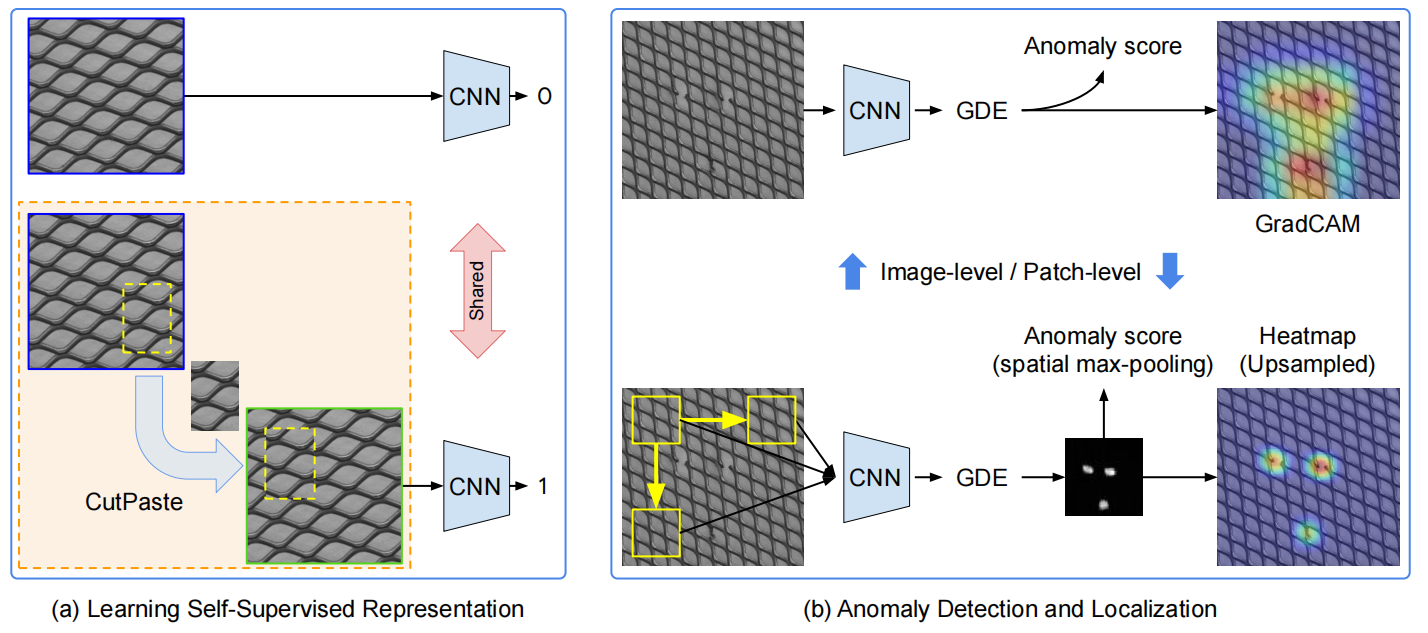}
\caption{Overview of CutPaste \cite{li2021cutpaste}. (a) The CutPaste method creates an abnormal image by cropping a random area of the defect-free image and superimposing it on the normal image. Then the CNN classifier is trained jointly with normal samples and synthetic anomalous samples. (b, top) Image-level inference process. Gaussian density estimator(GDE) is adopted to compute anomaly scores via one-class classifiers. (b, bottom) Patch-level inference process. Features are extracted from local patches to produce anomaly score map, which is then max-pooled for detection or upsampled for localization. }
\label{fig:cutpaste}
\end{figure*}

In DRAEM \cite{zavrtanik2021draem} method, noise generation method is adopted to create anomalies and superimpose them on normal images. The proposed method learns a joint representation of a normal image and its synthetic anomalous image, while simultaneously learns a decision boundary between normal and anomalous examples.

The method NSA \cite{schluter2021self} is a naturally synthetic anomaly approach that proposes a way to create anomalies by selecting patches of different sizes at different locations and blending them into anomaly-free images. Specifically, it is a self-supervised task to create diverse and realistic synthetic anomalies with Poisson image editing to seamlessly blend multiscale patches of various sizes in different images. This produces a wide range of synthetic anomalies, which are more similar to natural sub-image irregularities.

CutPaste \cite{li2021cutpaste} is also an synthetic anomaly method designed to produce augmentations to synthesize anomalous samples by operating on normal image patches, including cropping, rotating, transforming and overlaying. The distance between the normal samples and the generated anomalous samples is then measured. An overview of CutPaste method for anomaly detection and localization is shown in Fig.~\ref{fig:cutpaste}.

The method \cite{ristea2021self} proposes a self-supervised predictive convolutional attentive block (SSPCAB), which can be easily incorporated into various state-of-the-art anomaly detection methods, such as DRAEM \cite{zavrtanik2021draem} and CutPaste \cite{li2021cutpaste}. It aims at reconstructing masked information with contextual information, so as to realize performance improvements.

AnoSeg \cite{song2021anoseg} is a segmentation model which combines three techniques: self-supervised learning with hard augmentation, adversarial learning, and coordinate channel connectivity. It is directly trained for anomaly segmentation tasks with synthetic anomaly data generated by hard augmentation. In addition, anomaly regions sensitive to positional relationships are more easily to be detected by means of coordinate vectors representing the pixel position information.

\subsection{Algorithm enhancements}
 \label{section:enhancements}
Some algorithms provide some enhancements \cite{IGD,yi2020patch,liznerski2020explainable,reiss2021mean}, such as improved loss functions or interpretability.

IGD \cite{IGD} employs reverse-interpolated training samples to train a class of Gaussian anomaly classifiers that describe representative normal samples for effective normality. Current state-of-the-art models learn a compact normality description by hyper-sphere minimization, but they are prone to overfitting. To solve this problem, interpolated Gaussian descriptor (IGD) approach is introduced. Methods that can locate anomalies generally are suitable for a specific anomaly size and structure, which may result in missing anomalies outside of that size and structure range. To avoid this problem, IGD is designed to detect multiscale structural and non-structural anomalies to improve the accuracy of anomaly localization.

Classical unsupervised anomaly detection algorithms such as support vector
data description (SVDD \cite{tax2004support}) and Deep-SVDD (DSVDD \cite{ruff2018deep}) can hardly explain why an image is anomalous. Therefore, FCDD \cite{liznerski2020explainable} explores converting the final comparison vector of the previous DSVDD model into a two-dimensional matrix (explanation heatmap) to enhance the interpretability of the algorithm. For most traditional fully connected convolutional networks, images are mapped to the feature map of $1 * U * V$. It is mentioned in this paper that an important attribute of the convolution layer is that a pixel of the feature map only has a fixed receptive field corresponding to the input. A heatmap upsampling algorithm is proposed in this paper, so that the abnormal score of the feature map can be mapped back to the position of the original image, i.e., spatial information is reserved. 

A new loss function is proposed which can overcome failure modes of both center-loss and contrastive-loss methods \cite{reiss2021mean}. Furthermore, it is combined with a confidence-invariant center loss, which replaces the Euclidean distance used in previous work, i.e., a distance that is sensitive to prediction confidence. The improvements yield a new anomaly detection approach, based on mean-shifted contrastive loss, which is both more accurate and less sensitive to catastrophic model collapse than previous methods.

In the field of anomaly detection, attention mechanisms \cite{pirnay2021inpainting,venkataramanan2020attention} are often used for algorithm improvement. Another kind of methodology utilizes multiscale features to enrich semantic information capture \cite{rudolph2022fully,gudovskiy2022cflow,defard2021padim,rippel2021modeling,rudolph2021same,cohen2020sub,STPM,massoli2020mocca,yang2020dfr}.

There are other methods try brand new ways to solve anomaly detection task. For the first time, RFS Energy algorithm \cite{kamoona2021anomaly} solves the challenge of unsupervised anomaly detection using keypoint detection and an energy model.

\section{Comparison and Analysis}
\label{section:Analysis}

\begin{table}[]
\centering
\caption{Comparison between supervised and unsupervised algorithms.}
\label{tab:comparison}
\resizebox{0.45\textwidth}{!}{%
\begin{tabular}{@{}cccccc@{}}
\hline
& \begin{tabular}[c]{@{}c@{}}Need \\ label\end{tabular} & \begin{tabular}[c]{@{}c@{}}Support for \\ undefined \\defects\end{tabular} & Accuracy             & \begin{tabular}[c]{@{}c@{}}Generalizability\end{tabular} \\ \hline
Supervised   & \Checkmark  & \XSolidBrush      & higher         & \XSolidBrush    \\
~\ref{section:Reconstruction} Reconstruction-based & \XSolidBrush  & \Checkmark       & high        & \Checkmark    \\
~\ref{section:nf} NF-based   & \XSolidBrush  & \Checkmark      & high         & \Checkmark    \\
~\ref{section:Representation} Representation-based    & \XSolidBrush  & \Checkmark      & high         & \XSolidBrush \\
~\ref{section:augmentation} Augmentation-based  & \XSolidBrush  & \XSolidBrush      & low    & \XSolidBrush    \\
\hline
\end{tabular}
}
\end{table}

\begin{table}[]
\centering
\caption{Complexity comparison in terms of inference speed (FPS), additional inference time (millisecond) and number of additional parameters (M) for various backbones. A.d. Time means the additional inference time and A.d. Parmas is the number of additional parameters compared with backbone network.}
\label{tab:complex}
\begin{tabular}{@{}cccc@{}}
\hline
Model & FPS & A.d. Time (ms) & A.d. Params (M) \\ \hline
\begin{tabular}[c]{@{}c@{}}Patchcore \cite{roth2021towards}\\ (Representation-based)\end{tabular} & 5.88 & 159  & 0    \\
\begin{tabular}[c]{@{}c@{}}CFlow \cite{gudovskiy2022cflow}\\ (NF-based)\end{tabular}                 & 14.9 & 56   & 81.6 \\
\begin{tabular}[c]{@{}c@{}}DFR \cite{yang2020dfr}\\ (Reconstruction-based)\end{tabular}       & 100  & 43.8 & 124  \\ \hline
\end{tabular}
\end{table}

Both supervised and unsupervised algorithms are used in the field of anomaly detection, and the advantages and disadvantages of each are summarized in Table~\ref{tab:comparison}. Although the supervised approach possesses high accuracy, there are limitations in the acquisition of labeled data, which requires a large amount of work, and sometimes it is impossible to acquire enough labeled defect samples. The process of training the network also has many parameters to optimize, which leads to inefficiency. Classification is not possible for defects that do not appear in the training set. The classes of methods introduced above in this paper are all unsupervised algorithms that do not require category labels, which can save a lot of cost and effort in practical applications. NF-based methods require expensive training computational resources, while undefined defect detection is supported and inference efficiency is high. Reconstruction-based methods require expensive training for the related task and deep generative models are not robust enough, and their performances for anomaly detection are not stable whereas the model has good generalization ability. Representation-based methods do not need to introduce parameters other than backbone, which is beneficial for efficiency. However, because backbone is usually biased towards ImageNet, it does not have good generalizability for some images, such as medical images. Data augmentation-based 
methods are designed to resemble the anomalies, which are data-dependent and non-automatic.

As far as complexity is concerned, we take time complexity and memory complexity into account. \textbf{Time complexity.} For representation-based algorithms, the training time complexity scales linearly with the dataset size. However, contrary to the methods that require training deep neural networks like reconstruction-based methods, representation-based algorithms use a pre-trained CNN, and, thus, no deep learning training is required which is often a complex procedure. Hence, it is very fast and easy to train on small datasets like MVTec AD. Conversely, take SPADE as an example, it computes and stores in the memory before testing all the embedding vectors of the normal training images. Those vectors are the inputs of a KNN algorithm which makes SPADE’s inference speed very slow. While for reconstruction-based methods, after training stage, their inference phase can be quite fast. NF-based methods avoid the time-consuming k-nearest-neighbor-search process, while it still needs to perform a more complex inference phase than reconstruction-based methods. \textbf{Memory complexity.}
Representation-based algorithms like SPADE and Patchcore perform KNN clustering between each test feature of each image patch and the gallery features of normal image patches, and they do not need to introduce parameters other than backbone. But they require large memory allocation for gallery features.

We make an efficiency analysis of some representative methods from aspects of inference speed, additional inference time and additional model parameters, “additional” refers to not considering the backbone itself. The hardware configuration of the machine used for testing is Intel(R) Xeon(R) CPU E5-2680 V4@2.4GHZ and NVIDIA GeForce GTX 1080Ti. The analysis results are shown in Table~\ref{tab:complex}.

\section{Datasets}
 \label{section:Datasets}

Datasets are the base for research work. A good dataset is more conducive to the discovery and summary of problems, so as to facilitate the solution.
There are now some quality inspection/anomaly detection datasets in the industry field.

\subsection{BTAD}
BeanTech Anomaly Detection dataset \cite{MishraVFPF21} (BTAD\footnote{http://avires.dimi.uniud.it/papers/btad/btad.zip}) contains a total of 2830 real-world images of 3 industrial products showcasing body and surface defects. The training set consists of only normal images, while the testing set has a mixture of both normal and abnormal images. Product 0, 1, and 2 of this dataset contain 400, 1000, and 399 training images respectively. This dataset is often used for unsupervised defect/anomaly detection. The AUROC (area under the receiver operator curve) metrics of the SOTA methods on this dataset are summarized in Table~\ref{BATD}, where the bold parts are the best-performing results.

\begin{table}[]
\centering
\caption{Anomaly localization results measured by pixel-wise AUROC on BTAD dataset.}
\label{BATD}
\begin{tabular}{@{}ccccc@{}}
\hline
Methods     & Product0       & Product1       & Product2       & Average        \\ \hline
VT-ADL \cite{MishraVFPF21}     & \textbf{0.990} & 0.940          & 0.770          & 0.900          \\
AE MSE \cite{hinton2006reducing}     & 0.490          & 0.920          & 0.950          & 0.787          \\
AE MSE+SSIM \cite{BergmannLFSS19} & 0.530          & 0.960          & 0.890          & 0.793          \\
FastFlow \cite{yu2021fastflow}   & 0.950          & 0.960          & 0.990          & 0.967          \\
BGAD \cite{Explicit}        & 0.972          & 0.967          & 0.996          & 0.978          \\
BGAD-FAS \cite{Explicit}   & 0.980          & \textbf{0.977} & \textbf{0.998} & \textbf{0.985} \\
FYD \cite{zheng2021focus}        & 0.961          & 0.953          & 0.997          & 0.970          \\ \hline
\end{tabular}
\end{table}

\subsection{Solar panel dataset: ELPV}
The dataset ELPV\footnote{https://github.com/zae-bayern/elpv-dataset} \cite{Deitsch2019,Buerhop2018,Deitsch2021} contains 2624 8-bit grayscale image samples of 300 x 300 pixel functional and defective solar cells, with varying degrees of degradation extracted from 44 different solar modules. Defects in annotated images are internal or external types of defects known to reduce the power efficiency of solar modules. With every image annotated with a defect probability (a floating point value between 0 and 1), this dataset can be used to solve unsupervised tasks.

\subsection{Fabric defect dataset: AITEX}
The collection AITEX\footnote{http://www.aitex.es/afid/} \cite{silvestre2019public} contains photos of seven different fabric textures with a resolution of 4096 × 256 pixels. There are 140 defect-free images in the dataset, 20 images for each type of fabric. In addition, there are 105 images of 12 different types of fabric defects commonly found in the textile industry. It can be used to solve unsupervised tasks. The AUROC metrics of the SOTA methods on ELPV dataset and AITEX dataset are summarized in Table~\ref{AITEX}, where the bold parts are the best-performing results.
\begin{table}[]
\centering
\caption{Image-level anomaly detection AUROC results on the AITEX and ELPV datasets.}
\label{AITEX}
\begin{tabular}{@{}ccc@{}}
\hline
Methods  & AITEX          & ELPV           \\ \hline
KDAD \cite{SalehiSBRR21}     & 0.576          & 0.744          \\
DevNet \cite{Deviation}   & 0.598          & 0.514          \\
FLOS \cite{LinGGHD20}    & 0.538          & 0.457          \\
SAOE \cite{TackMJS20}    & 0.675          & 0.635          \\
MLEP \cite{liu2019margin}     & 0.564          & 0.578          \\
DRA \cite{ding2022catching}     & 0.692          & 0.675          \\
BGAD-FAS \cite{Explicit} & \textbf{0.826} & \textbf{0.903} \\ \hline
\end{tabular}
\end{table}

\subsection{MTD-Surface defect saliency}
In magnetic brick surface defect dataset\footnote{https://github.com/abin24/Magnetic-tile-defect-datasets.} \cite{huang2020surface}, a total of 1344 images are taken, the ROI (region of interest) of the tiles is cropped and classified into six subsets according to defect type, which are respectively porosity, crack, wear, fracture, non-uniformity (caused by the grinding process) and free (defect-free), each with a pixel-level label. To simulate the manufacturing process on an actual assembly line, images are captured under a variety of lighting conditions for each given brick. It can be used to solve unsupervised tasks. The experiment results of AUROC on this dataset are summarized in Table~\ref{MTD}, where the bold parts are the best-performing results.
\begin{table}[]
\centering
\caption{Experiment results of AUROC on MTD dataset.}
\label{MTD}
\begin{tabular}{@{}cc@{}}
\hline
Methods     & AUROC          \\ \hline
Geom \cite{golan2018deep}        & 0.755          \\
GANomaly \cite{akcay2018ganomaly}    & 0.766          \\
DSEBM \cite{zhai2016deep}      & 0.572          \\
Mahalanobis \cite{rippel2021modeling}& 0.980          \\
1-NN \cite{nazare2018pre}       & 0.800          \\
DifferNet \cite{rudolph2021same}   & 0.977          \\
PaDiM \cite{defard2021padim}      & 0.987          \\
CS-Flow \cite{rudolph2022fully}    & \textbf{0.993} \\
PatchCore \cite{roth2021towards}  & 0.979          \\
ADGAN \cite{cheng2020adgan}      & 0.464          \\
OCSVM \cite{andrews2016transfer}      & 0.587          \\ \hline
\end{tabular}
\end{table}
\subsection{KolektorSDD}
KolektorSDD \cite{KolektorSDD} consists of 399 images of electrical commutators, where 52 defected images are annotated for microscopic fractions or cracks on the surface of the plastic embedding in electrical commutators. The dataset represents a real-world problem of surface-defect detection for an industrial semi-finished product where the number of defective items available for the training is limited. Table~\ref{KolektorSDD} shows AUROC performance on KolektorSDD dataset in terms of several SOTA algorithms.

\begin{table}[]
\centering
\caption{AUROC performance on KolektorSDD dataset.}
\label{KolektorSDD}
\begin{tabular}{@{}cc@{}}
\hline
Methods            & AUROC          \\ \hline
skipGAN \cite{akccay2019skip}           & 0.551          \\
Puzzle AE \cite{salehi2020puzzle}         & 0.554          \\
DifferNet \cite{rudolph2021same}         & 0.849          \\
InTra \cite{pirnay2021inpainting}             & 0.701          \\
CutPaste \cite{li2021cutpaste}          & 0.602          \\
Draem \cite{zavrtanik2021draem}             & 0.859          \\
OCR-GAN \cite{liang2022omni}           & 0.914          \\
Uninformed student \cite{bergmann2020uninformed}& 0.896          \\
PaDiM \cite{defard2021padim}             & 0.945          \\
Semi-orthogonal \cite{kim2021semi}   & \textbf{0.960} \\ \hline
\end{tabular}
\end{table}
\begin{table}[]
\centering
\caption{AUROC performance on DAGM dataset.}
\label{DAGM}
\begin{tabular}{@{}cc@{}}
\hline
Methods      & Average        \\ \hline
skipGAN \cite{akccay2019skip}     & 0.558          \\
Puzzle AE \cite{salehi2020puzzle}   & 0.593          \\
CutPaste \cite{li2021cutpaste}    & 0.665          \\
DifferNet \cite{rudolph2021same}    & 0.746          \\
Draem \cite{zavrtanik2021draem}       & 0.980          \\
OCR-GAN \cite{liang2022omni}     & \textbf{0.993} \\
f-AnoGAN \cite{schlegl2019f}    & 0.575          \\
Uninformed student \cite{bergmann2020uninformed} & 0.864          \\
Staar \cite{staar2019anomaly}       & 0.830          \\ \hline
\end{tabular}
\end{table}
\begin{table*}[]
\centering
\caption{Anomaly detection and localization performance on MVTec AD dataset with the format (image-level AUROC, pixel-level AUROC).}
\label{MVTec}
\resizebox{\textwidth}{!}{%
\begin{tabular}{@{}ccccccccc@{}}
\hline
 &
  PatchSVDD \cite{yi2020patch} &
  SPADE \cite{cohen2020sub} &
  CutPaste \cite{li2021cutpaste} &
  PatchCore \cite{roth2021towards} &
  \multicolumn{1}{l}{AST \cite{rudolph2023asymmetric}} &
  CFA \cite{lee2022cfa} &
  CFlow \cite{gudovskiy2022cflow} &
  FastFlow \cite{yu2021fastflow} \\ \hline
carpet     & (92.9,92.6)  & (98.6,97.5) & (100.0,98.3) & (98.7,98.9)  & (97.5,-)    & (97.3,-)    & (100.0,99.3) & (100.0,99.4) \\
grid       & (94.6,96.2)  & (99.0,93.7) & (96.2,97.5)  & (98.2,98.7)  & (99.1,-)    & (99.2,-)    & (97.6,99.0)  & (99.7,98.3)  \\
leather    & (90.9,97.4)  & (99.5,97.6) & (95.4,99.5)  & (100.0,99.3) & (100.0,-)   & (100.0,-)   & (97.7,99.7)  & (100.0,99.5) \\
tile       & (97.8,91.4)  & (89.8,87.4) & (100.0,90.5) & (98.7,95.6)  & (100.0,-)   & (99.4,-)    & (98.7,98.0)  & (100.0,96.3) \\
wood       & (96.5,90.8)  & (95.8,88.5) & (99.1,95.5)  & (99.2,95.0)  & (100.0,-)   & (99.7,-)    & (99.6,96.7)  & (100.0,97.0) \\
bottle     & (98.6,98.1)  & (98.1,98.4) & (99.9,97.6)  & (100.0,98.6) & (100.0,-)   & (100.0,-)   & (100.0,99.0) & (100.0,97.7) \\
cable      & (90.3,96.8)  & (93.2,97.2) & (100.0,90.0) & (99.5,98.4)  & (98.5,-)    & (99.8,-)    & (100.0,97.6) & (100.0,98.4) \\
capsule    & (76.7,95.8)  & (98.6,99.0) & (98.6,97.4)  & (98.1,98.8)  & (99.7,-)    & (97.3,-)    & (99.3,99.0)  & (100.0,99.1) \\
hazelnut   & (92.0,97.5)  & (98.9,99.1) & (93.3 ,97.3) & (100.0.98.7) & (100.0,-)   & (100.0,-)   & (96.8,98.9)  & (100.0,99.1) \\
meta nut   & (94.0,98.0)  & (96.9,98.1) & (86.6,93.1)  & (100.0,98.4) & (98.5,-)    & (100.0,-)   & (91.9,98.6)  & (100.0,98.5) \\
pill       & (86.1,95.1)  & (96.5,96.5) & (99.8,95.7)  & (96.6,97.1)  & (99.1,-)    & (97.9,-)    & (99.9,99.0)  & (99.4,99.2)  \\
screw      & (81.3,95.7)  & (99.5,98.9) & (90.7,96.7)  & (98.1,99.4)  & (99.7,-)    & (97.3,-)    & (99.7,98.9)  & (97.8,99.4)  \\
toothbrush & (100.0,98.1) & (98.9,97.9) & (97.5 ,98.1) & (100.0,98.7) & (96.6,-)    & (100.0,-)   & (95.2,99.0)  & (94.4,98.9)  \\
transistor & (91.5,97.0)  & (81.0,94.1) & (99.8,93.0)  & (100.0,96.3) & (99.3,-)    & (100.0,-)   & (99.1,98.0)  & (99.8,97.3)  \\
zipper     & (97.9,95.1)  & (98.8,96.5) & (99.9,99.3)  & (98.8,98.8)  & (99.1,-)    & (99.6,-)    & (98.5,99.1)  & (99.5,98.7)  \\\hline
Average    & (92.1,95.7)  & (96.2,96.5) & (97.1,96.0)  & (99.1,98.1)  & (99.2,95.0) & (99.2,98.2) & (98.3,98.6)  & (99.4,98.5)  \\ \hline
\end{tabular}%
}
\end{table*}
\begin{figure*}[h]%
\centering
\includegraphics[width=1.0\textwidth]{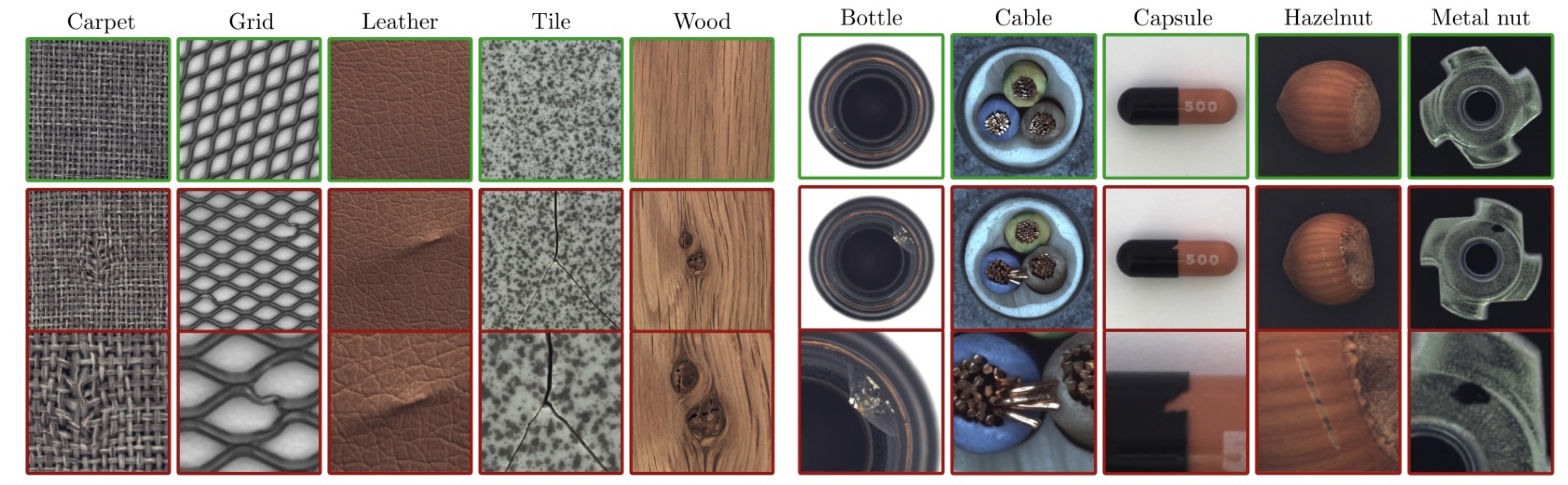}
\caption{Example images of the MVTec AD dataset. For each category, the top row shows an anomaly-free image. The middle row shows an anomalous example. In the bottom row, a close-up view that highlights the anomalous region is provided.}\label{fig:mvtec}
\end{figure*}
\begin{table*}[ht]
\centering
\caption{Challenges in anomaly detection. Different datasets illustrate the challenges in the industry anomaly detection field.}
\label{tab:tickingbox}
\resizebox{0.8\textwidth}{!}{%
\begin{tabular}{@{}cccccccc@{}}
\hline
                               & BTAD & ELPV & AITEX & MTD-Surface & KolektorSDD & DAGM & MVTec AD \\ \hline
Small amount of anomalous data &      &      & \Checkmark     &             & \Checkmark           &      &          \\
Small size of defects          &      &      & \Checkmark     &             &             & \Checkmark    &          \\
Object appearance variability  & \Checkmark    & \Checkmark    &       &             &             &      & \Checkmark        \\
Texture differences            &      &      & \Checkmark     &\Checkmark           &             &      &          \\ \hline
\end{tabular}%
}
\end{table*}

\subsection{DAGM}
DAGM\footnote{https://conferences.mpi-inf.mpg.de/dagm/2007/prizes.html} \cite{wieler2007weakly} is a well-known benchmark dataset for surface defect detection. It contains images of various surfaces with artificially generated defects. Surfaces and defects are split into 10 classes of various difficulties, such as scratches or spots. It is a weakly supervised dataset, and there are 8,050 training and testing sets each, and the ratio of positive and negative samples for each type is approximately 1:7. The experiment results of AUROC on this dataset averaged over ten categories are summarized in Table~\ref{DAGM}, where the bold part 
is the best-performing result. 

\subsection{MVTec AD}
MVTec AD dataset\footnote{http://www.mvtec.com/company/research/datasets} \cite{bergmann2019mvtec} has a total of 15 categories, with 5 of them being distinct types of textures and the remaining 10 being different sorts of objects. In total, 3629 photos are utilized for training and verification, while 1725 images are used for testing in this dataset. The training set contains solely non-defective images, whereas the test set contains both non-defective and defective images of various types. This dataset is often used for unsupervised defect/anomaly detection. Example images of MVTec AD dataset are shown in Fig.~\ref{fig:mvtec}. Under the metrics of image-level AUROC and pixel-level AUROC, the detailed comparison results of all categories are shown
in Table~\ref{MVTec}.

\subsection{Summary}
There are many challenges in visual industrial anomaly detection scenarios. Take the datasets we listed for example, as shown in Table~\ref{tab:tickingbox}, there are problems such as small amount of anomalous data, small size of defects, object appearance variability, texture differences, etc.
\begin{figure*}[htbp]%
\centering
    \begin{minipage}[t]{0.8\linewidth} 
        \centering 
        \includegraphics[width =.32\linewidth]{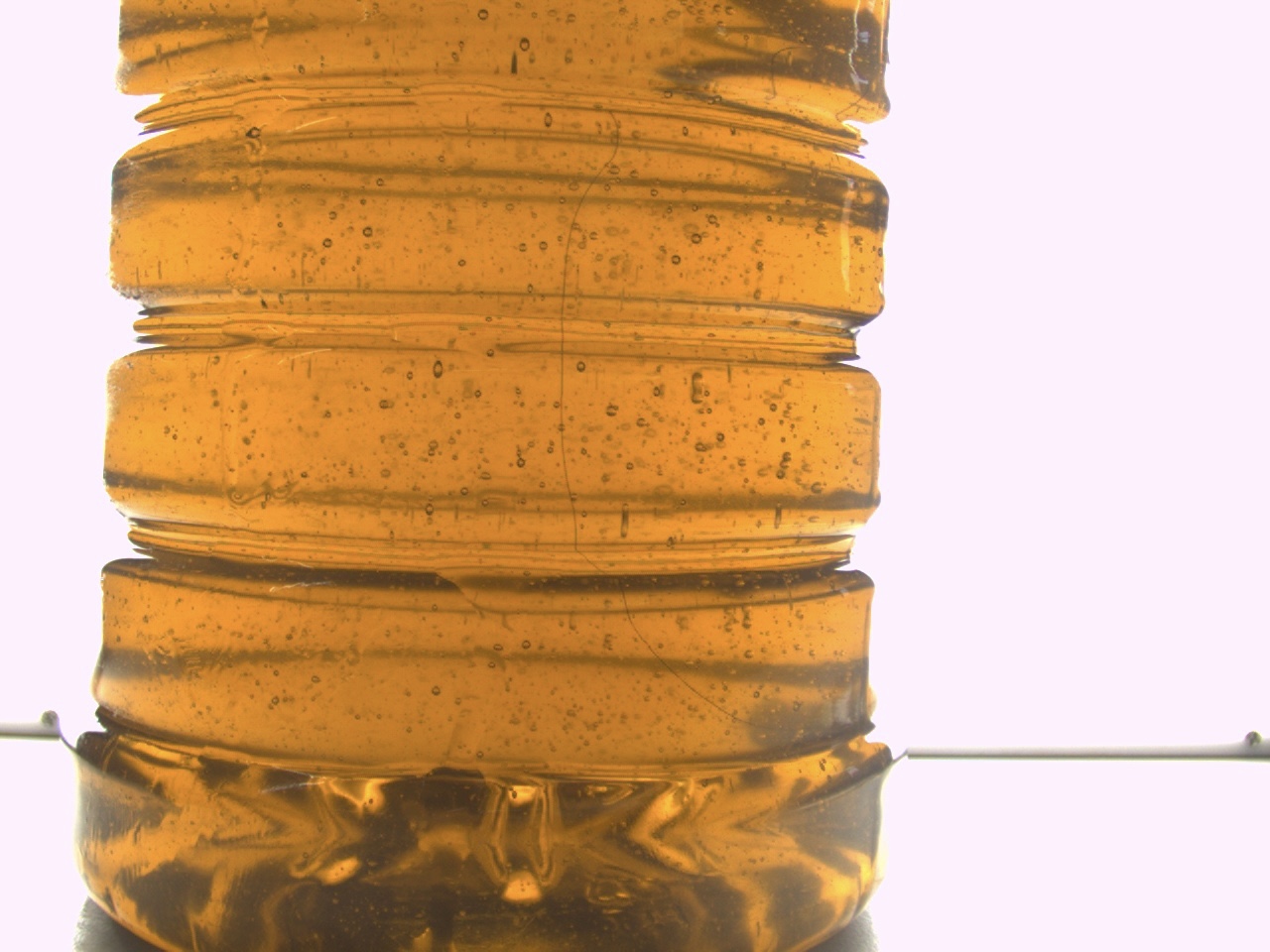}
        \includegraphics[width =.32\linewidth]{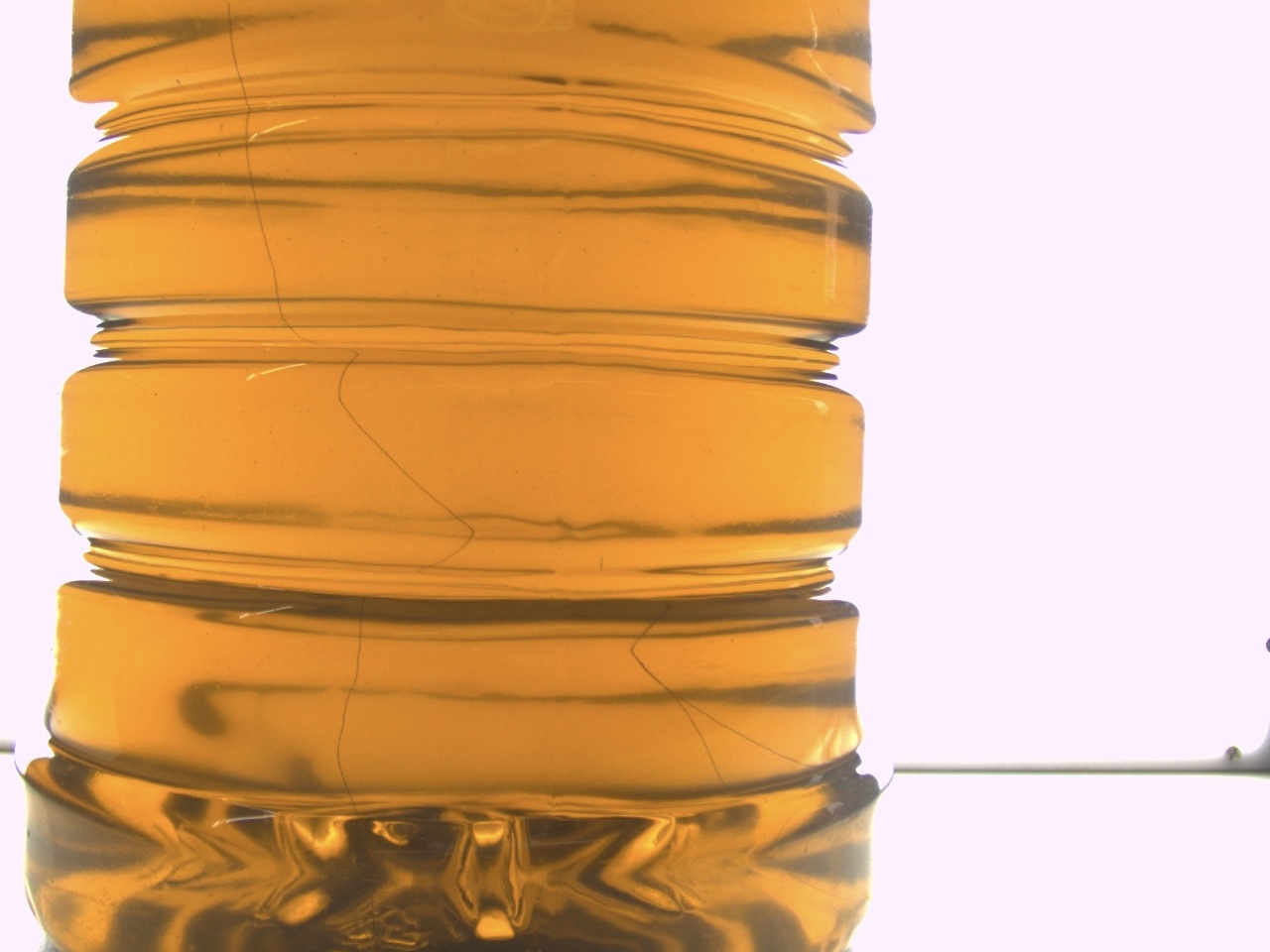}
        \includegraphics[width =.32\linewidth]{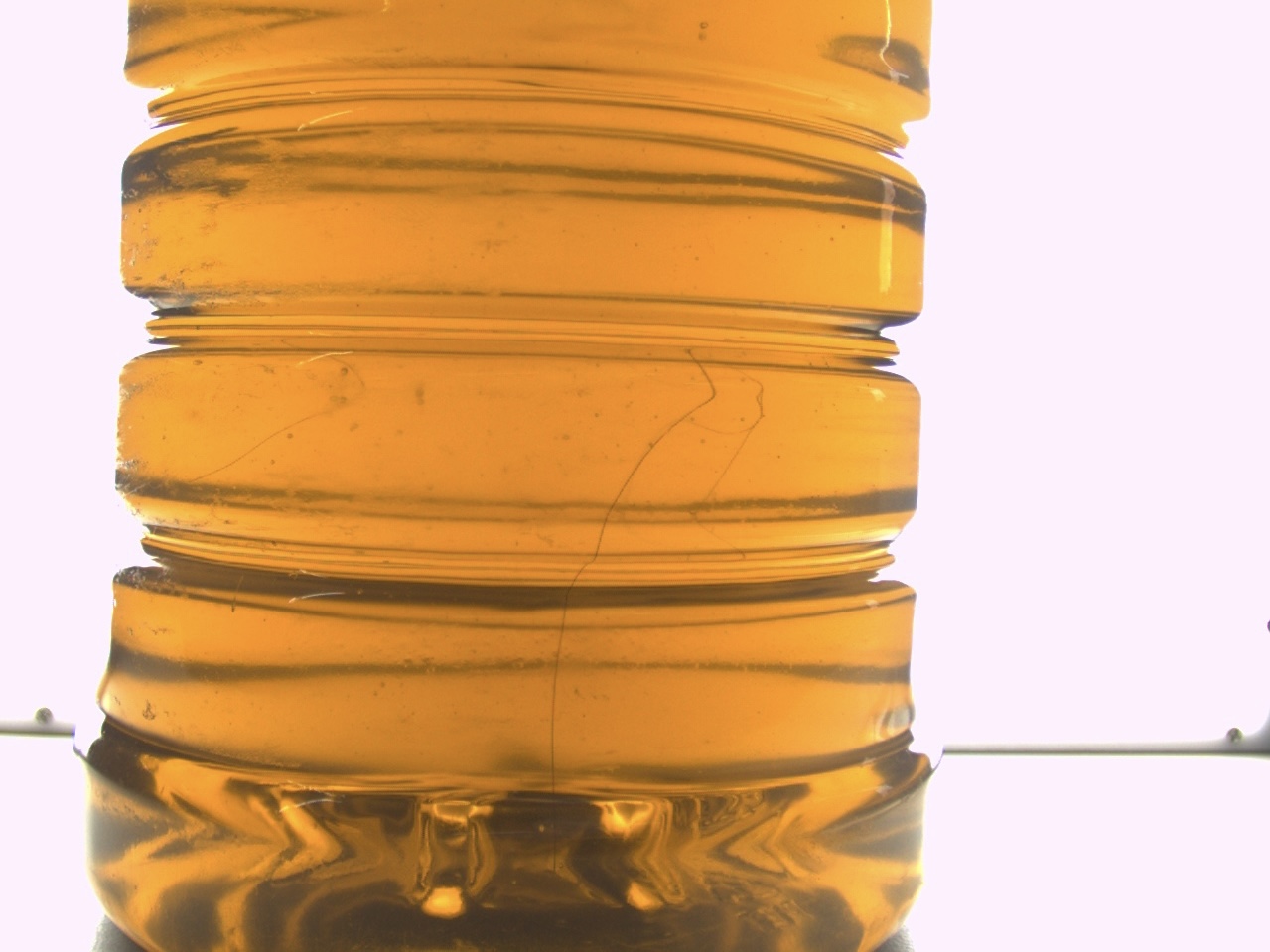}
    
        \subcaption{Edible oil impurity (hair)} 
    \end{minipage}%
    
    \begin{minipage}[t]{0.8\linewidth} 
        \centering 
        \includegraphics[width =.32\linewidth]{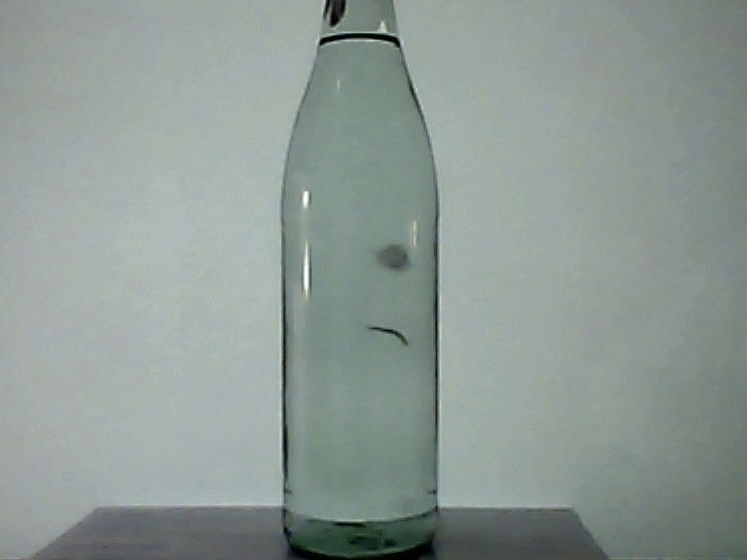}
        \includegraphics[width =.32\linewidth]{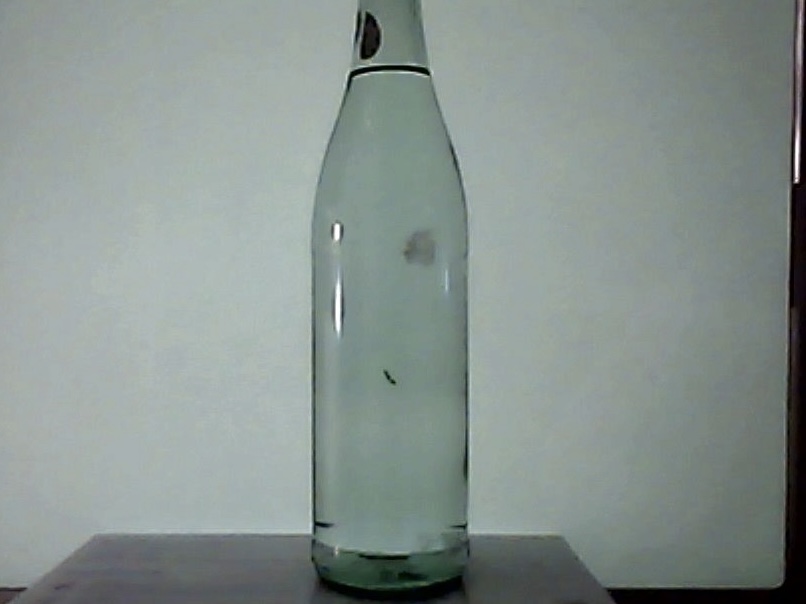}
        \includegraphics[width =.32\linewidth]{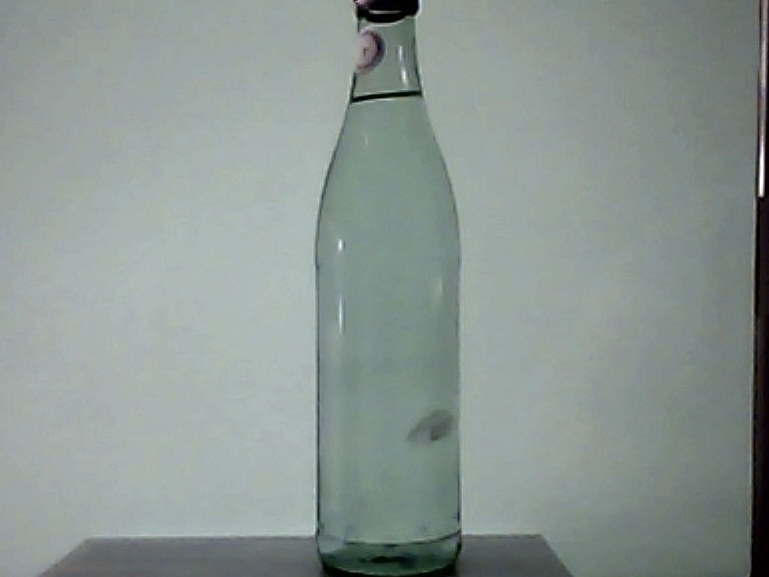}
    
        \subcaption{Wine impurity} 
    \end{minipage} 
    
    \begin{minipage}[t]{0.8\linewidth} 
        \centering 
        \includegraphics[width =.32\linewidth]{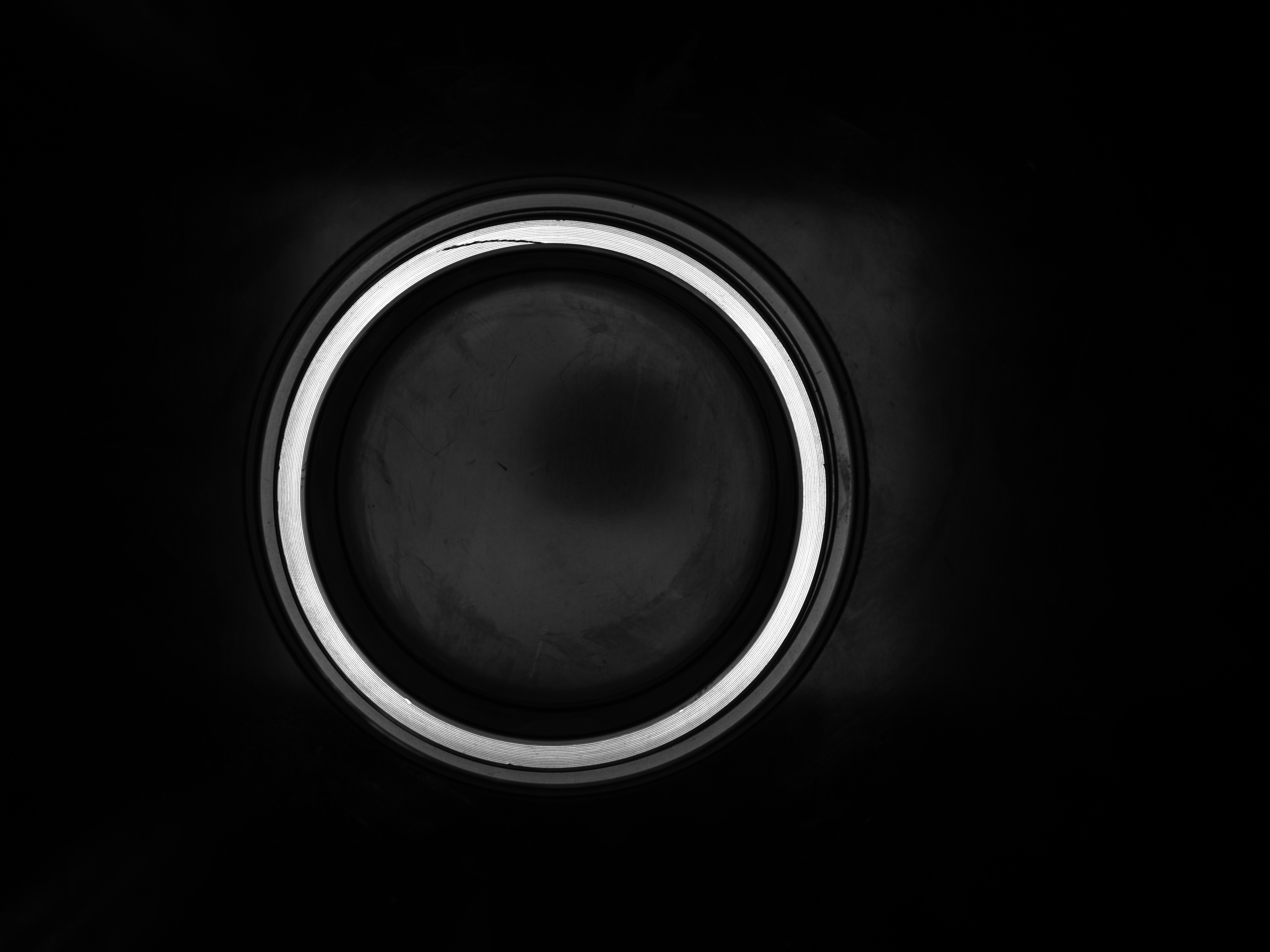}
        \includegraphics[width =.32\linewidth]{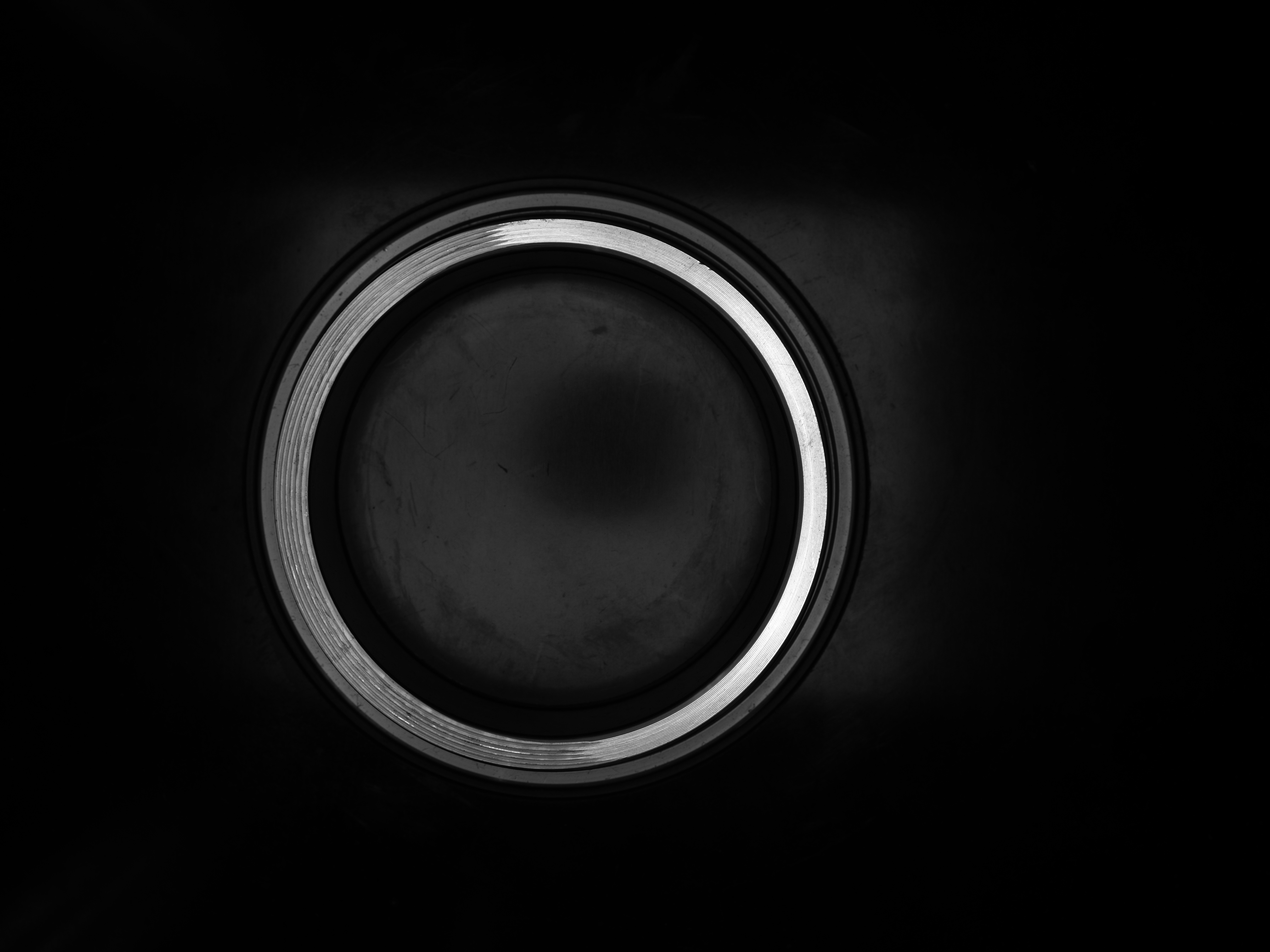}
        \includegraphics[width =.32\linewidth]{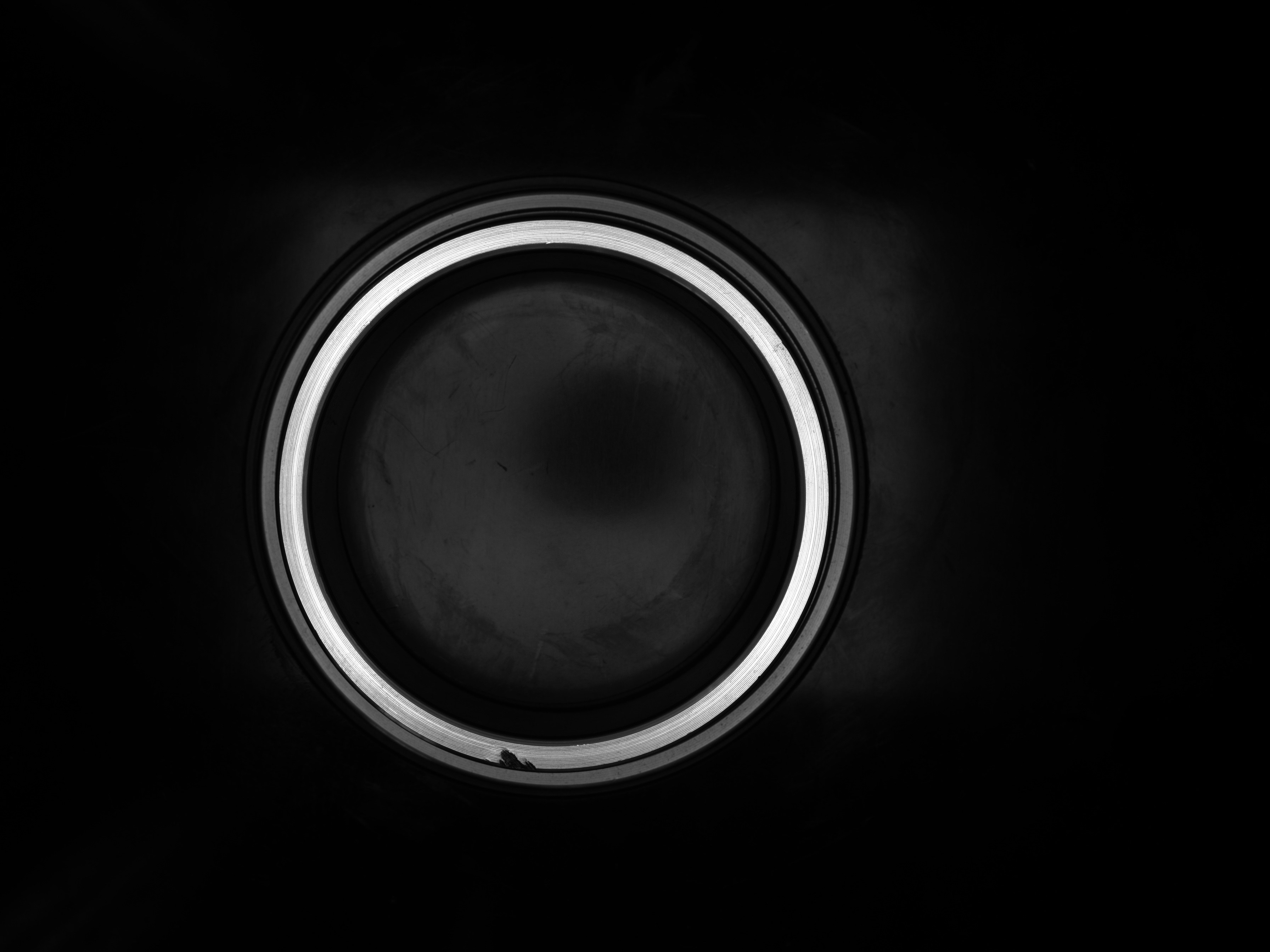}
        
        \subcaption{Bearing defects} 
    \end{minipage} 
    
    \begin{minipage}[t]{0.8\linewidth} 
        \centering 
        \includegraphics[width =.32\linewidth]{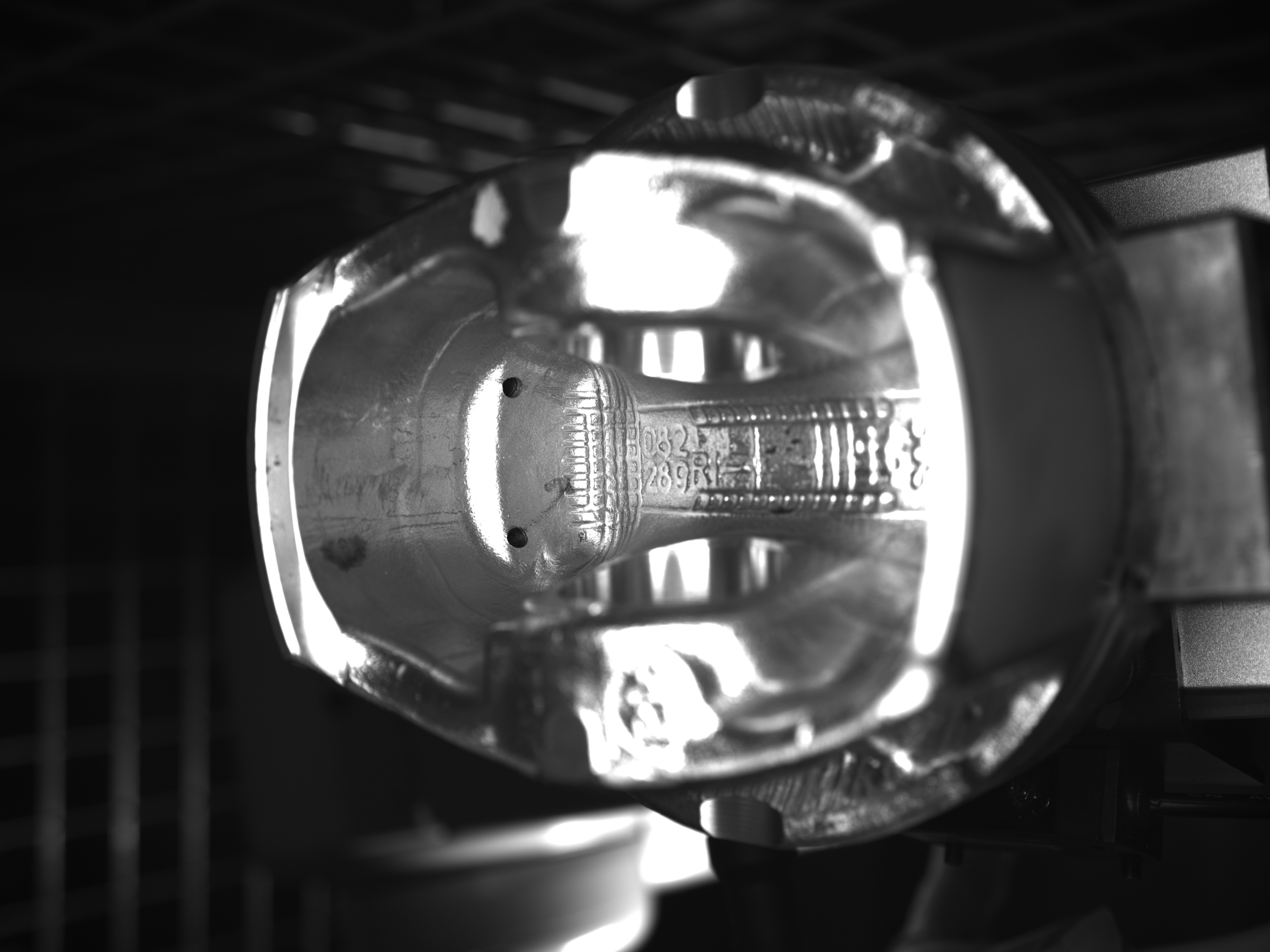}
        \includegraphics[width =.32\linewidth]{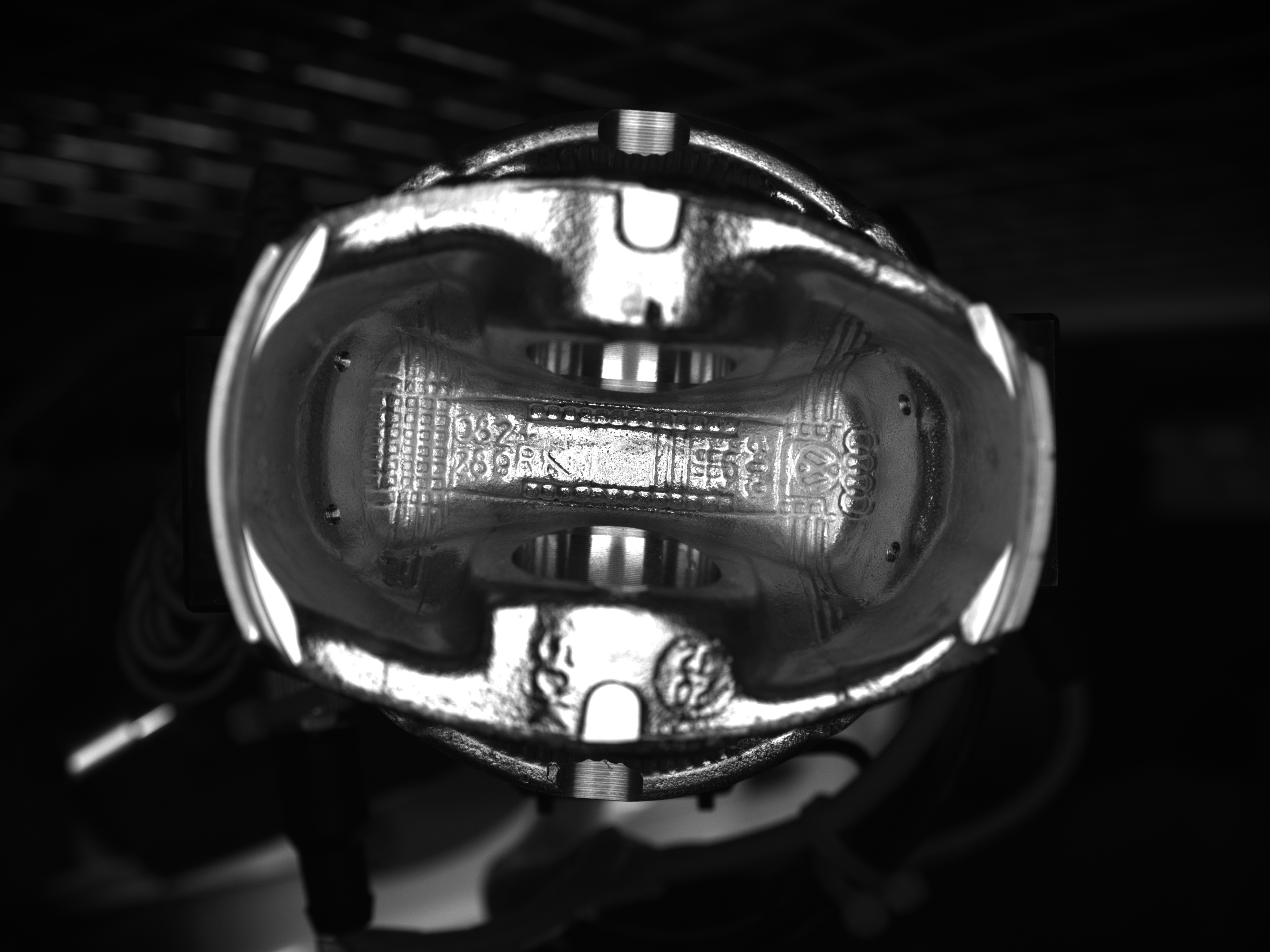}
        \includegraphics[width =.32\linewidth]{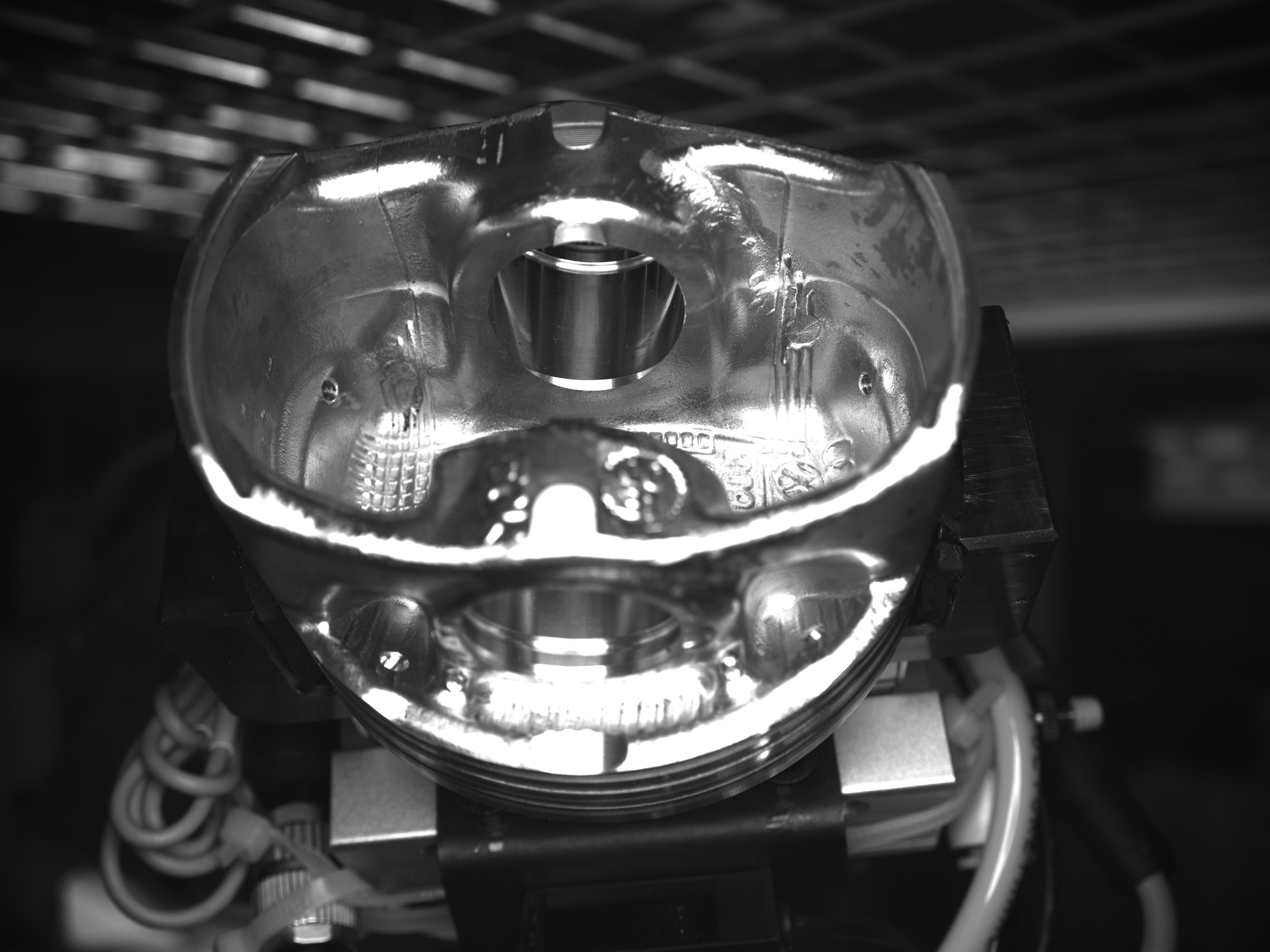}
    
        \subcaption{Engine lining defects} 
    \end{minipage} 

\caption{Examples of defects of different industrial materials.\label{fig:flawsample}}
\end{figure*}
\section{Challenges and Discussion}
\label{section:Challenges}
\textbf{Lack of comprehensive open datasets.} Currently, the existing open datasets merely cover a limited number of scenarios, which is not comprehensive enough. The actual industrial scenarios are rich and diverse, resulting in a domain gap with the scenarios presented by the open datasets. Although the AUROC of existing methods on open datasets is high, it is not sufficiently instructive. 
In industrial quality inspection scenarios, defects are complex and diverse, and the current study is only the tip of the iceberg. In the actual industrial scenario, there are also cases such as edible oil impurities, wine impurities, bearing defects, engine lining defects (3D internal), etc., as shown in Fig.~\ref{fig:flawsample}. There are still some problems that are not well solved by current methods.
Taking abnormal detection of edible oil in the actual industrial scene as an example, we verify the anomaly detection of image level with existing SOTA algorithm Patchcore and AST as shown in Table.~\ref{tab:oil-auc}, but the experimental results are not ideal and far from the results reported on MVTec benchmark. It further explains that the data in the actual industrial scenario is more complex, and the benchmark data is too simple and not rich enough. On the other hand, the lack of types and quantities of testing samples can not fully verify the proposed model is reliable, which hinders the generalization ability of the model. Therefore, it is necessary to launch richer datasets with diverse scenarios and testing samples.

\begin{table}[]
\centering
\caption{Experimental results of existing SOTA methods on edible oil data.}
\label{tab:oil-auc}
\resizebox{0.45\textwidth}{!}{%
\begin{tabular}{cccc}
\hline
                     & Patchcore-single     & Patchcore-3model     & AST                  \\ \hline
detection AUROC      & 92.2                 & 94.4                 & 91.8                 \\
\hline
\end{tabular}%
}
\end{table}

\noindent\textbf{Conflict between FAR and MAR.} In industrial applications there are intractably practical problems, FAR and MAR being a pair of contradictions. 
Correspondingly, the algorithm should be optimized to achieve a reduction in both false alarm rate and missed alarm rate. Otherwise missed alarms can lead to the production of inferior products, which will cause commercial loss, whereas a high rate of false alarms can lead to increased costs for manual confirmation. 

\noindent\textbf{Combination of data distribution learning and data augmentation.} Normalizing flow (NF)-based methods transform a simple distribution, such as a Gaussian distribution, into a more complex distribution by applying a series of invertible transformations. For unsupervised anomaly detection, NF is used to learn the distribution of normal samples \cite{yu2021fastflow}. There are other approaches \cite{makhzani2015adversarial} also based on the idea of learning data distribution. Considering the idea of learning data distribution in combination with data augmentation, NF can also be used to learn the distribution of artificially augmented defective samples. The joint learning of normal and artificial anomaly samples is beneficial to improve the generalization ability of the model. 

\noindent\textbf{Further research on foundation model.} Represention-based methods apply a pre-trained model to extract image features for anomaly detection, which demonstrates the effectiveness of the pre-trained model. As data volume and model scale evolve, foundation model \cite{bommasani2021opportunities} \cite{he2022masked} shows great potential as a member of pre-trained models. Foundation models are trained on massive amounts of data, which enables them to capture a broad range of patterns and relationships. By fine-tuning the model on specific tasks, it can quickly adapt to new domains and produce high-quality representation. 
The foundation model has striking strengths in representing ability and adaptation efficiency, and it has been started to be utilized in the field of computer vision \cite{xi2022ufo} \cite{riquelme2021scaling}. While the relevant research in the field of industrial anomaly detection still needs further exploration.
\begin{figure}[htbp]%
\centering
\includegraphics[width=0.45\textwidth]{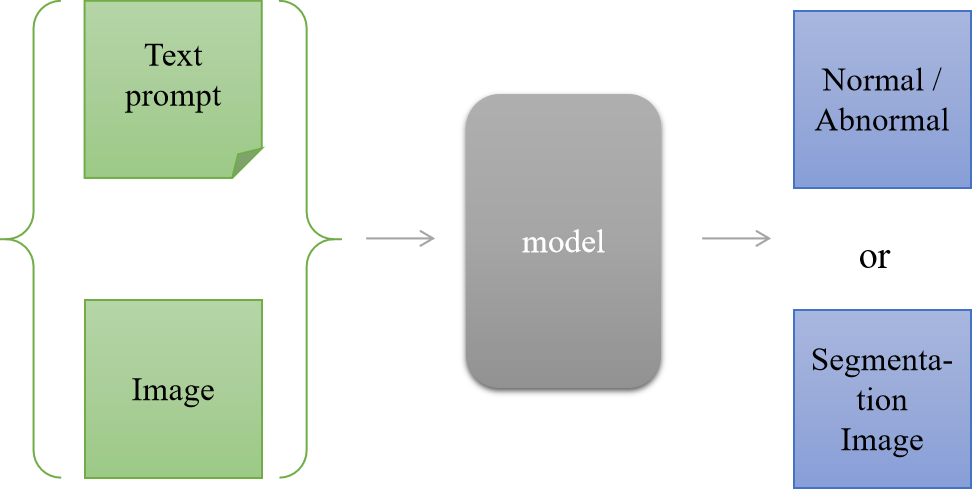}
\caption{Future model pipeline. Input image and text prompt, and the model can output classification or defect segmentation.}\label{future_model}
\end{figure}

\noindent\textbf{Multimodal industrial anomaly detection.} Multimodal learning can facilitate deeper understanding by providing multiple perspectives and facilitating connections between different modalities. With the development of multimodal learning, models have shown great potential in dealing with image and text modalities, like GPT-4 \cite{gpt4}, CLIP \cite{radford2021learning}, stable diffusion \cite{rombach2022high}, SAM \cite{kirillov2023segment}, OFA \cite{wang2022ofa} and Unified-IO \cite{lu2022unified}. In future research on industrial anomaly detection, it is expected to accept image and text prompt inputs and produce specified results, such as normal/abnormal classification or defect segmentation, as shown in Fig.~\ref{future_model}.

\section{Conclusion}
\label{section:Conclusion}
Deep learning has inspired a surge of interest in the visual industrial anomaly detection problem in recent years, resulting in a wide range of creative solutions. We present a complete review of newly proposed methodologies for visual industrial anomaly detection in the literature in this study. We categorize the relevant approaches based on their fundamental principles and describe their assumptions, benefits, and drawbacks, which may be of interest to practitioners as well as academic researchers. We hope to assist academics in better understanding the common principles of visual industrial anomaly detection systems and identifying interesting research directions in this area. 
The unsupervised anomaly detection algorithm is still under continuous research and development, and we will continue to track the progress in the follow-up work.
\begin{IEEEbiography}[{\includegraphics[width=1in,height=1.25in,clip,keepaspectratio]{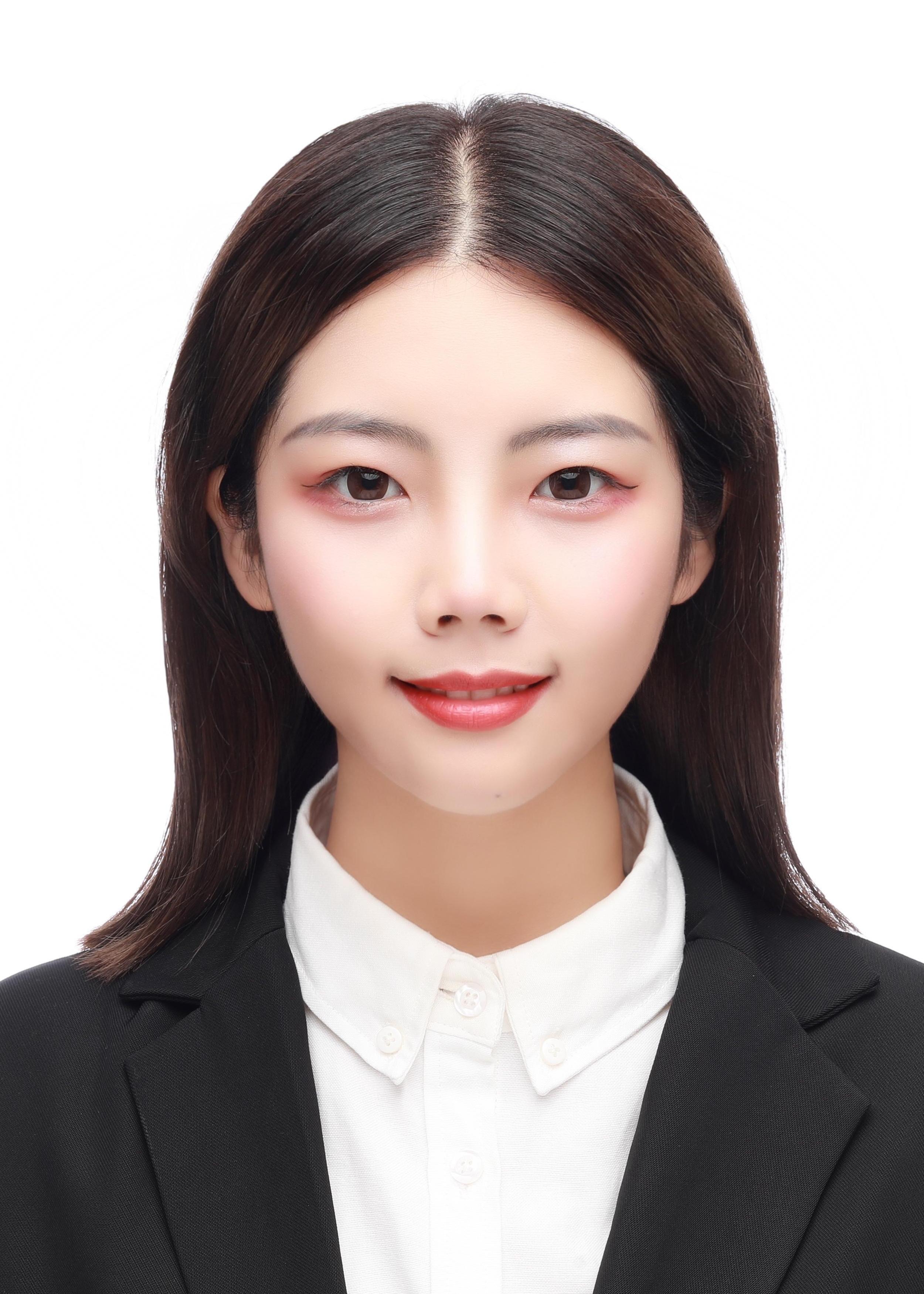}}]{YAJIE CUI}  received the B.S. degree in computer science and technology from Northeastern University, Shenyang, China, in 2018 and the M.S. degree in computer science and technology from Tianjin University, Tianjin, China, in 2021.

Since 2021, she has been an Deep Learning Algorithm Engineer of AI Innovation and Appliaction Center, China Unicom Digital Technology Co., Ltd, Beijing, China. Her research interest includes anomaly detection in computer vision, machine learning and deep learning.
\end{IEEEbiography}

\begin{IEEEbiography}[{\includegraphics[width=1in,height=1.25in,clip,keepaspectratio]{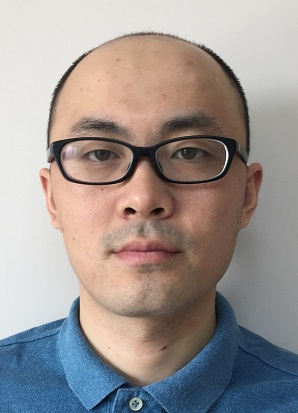}}]{ZHAOXIANG LIU} 
received the B.S. and Ph.D. degrees from the College of Information and Electrical Engineering, China Agricultural University, Beijing, China, in 2006 and 2011, respectively.

He joined VIA Technologies, Inc., Beijing, in 2011. From 2012 to 2016, he was a Senior Researcher of the Central Research Institute, Huawei Technologies, Beijing. He was a Senior Manager of CloudMinds Technologies Inc., Beijing, from 2016 to 2019. Since 2019, he has been a director of AI Innovation and Appliaction Center, China Unicom Digital Technology Co., Ltd, Beijing, China. His current research interests include computer vision, deep learning, robotics, and human-computer interaction.
\end{IEEEbiography}

\begin{IEEEbiography}[{\includegraphics[width=1in,height=1.25in,clip,keepaspectratio]{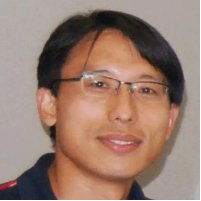}}]{SHIGUO LIAN} 
(M'04) received the Ph.D. degree from the Nanjing University of Science and Technology, China.

He was a Research Assistant with the City University of Hong Kong, Hong Kong, in 2004. From 2005 to 2010, he was a Research Scientist with France Telecom Research and Development Beijing, Beijing, China. He was a Senior Research Scientist and the Technical Director of the Huawei Central Research Institute, Beijing, from 2010 to 2016. He was a Senior Director of CloudMinds Technologies Inc., Beijing, from 2016 to 2019. Since 2019, he has been the Chief AI Scientist of China Unicom Digital Technology Co., Ltd and the General Manager of AI Innovation and Appliaction Center, China Unicom Digital Technology Co., Ltd, Beijing, China. He has authored over 80 refereed international journal articles covering topics of articial intelligence, multimedia communication, and human-computer interface. He has authored or coedited over ten books. He holds over 50 patents. 

Dr. Lian is on the editor board of several refereed international journals.
\end{IEEEbiography}

\EOD
\bibliography{sn-bibliography}
\bibliographystyle{IEEEtran}

\end{document}